\documentclass[10pt,twocolumn,letterpaper]{article}

\usepackage{cvpr}              % To produce the CAMERA-READY version
\usepackage{graphicx}
\usepackage{amsmath}
\usepackage{amssymb}
\usepackage{booktabs}

\usepackage{multirow}
\usepackage{bbding}
\usepackage{epsfig}

\usepackage{amsmath}
\usepackage{times}
\usepackage{epsfig}
\usepackage{graphicx}
\usepackage{amssymb}
\usepackage{bm}
\usepackage{amsfonts,amssymb}
\usepackage{float}

\usepackage{makecell}
\usepackage{multirow}
\usepackage{bm}
\usepackage[ruled]{algorithm2e}
\usepackage[accsupp]{axessibility}
\newlength\savewidth
\newcommand\shline{\noalign{\global\savewidth\arrayrulewidth
                            \global\arrayrulewidth 0.8pt}%
                   \hline
                   \noalign{\global\arrayrulewidth\savewidth}}

\usepackage[pagebackref,breaklinks,colorlinks]{hyperref}

\usepackage[capitalize]{cleveref}
\crefname{section}{Sec.}{Secs.}
\Crefname{section}{Section}{Sections}
\Crefname{table}{Table}{Tables}
\crefname{table}{Tab.}{Tabs.}

\def\cvprPaperID{4093} % *** Enter the CVPR Paper ID here
\def\confName{CVPR}
\def\confYear{2022}

\begin{document}

%%%%%%%%% TITLE - PLEASE UPDATE
\title{Deep Generalized Unfolding Networks for Image Restoration}

\author{
Chong Mou$^{\dagger}$, Qian Wang$^{\dagger}$, Jian Zhang$^{\dagger, \ddagger}$\\
$^\dagger$Peking University Shenzhen Graduate School, Shenzhen, China\\
$^\ddagger$Peng Cheng Laboratory, Shenzhen, China\\
{\tt\small eechongm@gmail.com; lucywq1028@gmail.com; zhangjian.sz@pku.edu.cn}
}
% \author{First Author\\
% Institution1\\
% Institution1 address\\
% {\tt\small firstauthor@i1.org}
% % For a paper whose authors are all at the same institution,
% % omit the following lines up until the closing ``}''.
% % Additional authors and addresses can be added with ``\and'',
% % just like the second author.
% % To save space, use either the email address or home page, not both
% \and
% Second Author\\
% Institution2\\
% First line of institution2 address\\
% {\tt\small secondauthor@i2.org}
% }

\maketitle
\let\thefootnote\relax\footnotetext{This work was supported in part by Shenzhen Fundamental Research Program (No.GXWD20201231165807007-20200807164903001) and National Natural Science Foundation of China (61902009).
% , and CAAI-Huawei Mindspore Open Fund~\cite{mindspore}. 
% We will implement our method via MindSpore in the future work. 
(\textit{Corresponding author: Jian Zhang.})}

%%%%%%%%% ABSTRACT
\begin{abstract}
Deep neural networks (DNN) have achieved great success in image restoration. However, most DNN methods are designed as a black box, lacking transparency and interpretability. Although some methods are proposed to combine traditional optimization algorithms with DNN, they usually demand pre-defined degradation processes or handcrafted assumptions, making it difficult to deal with complex and real-world applications. In this paper, we propose a \textbf{D}eep \textbf{G}eneralized \textbf{U}nfolding \textbf{Net}work (DGUNet) for image restoration. Concretely, without loss of interpretability, we integrate a gradient estimation strategy into the gradient descent step of the Proximal Gradient Descent (PGD) algorithm, driving it to deal with complex and real-world image degradation. In addition, we design inter-stage information pathways across proximal mapping in different PGD iterations to rectify the intrinsic information loss in most deep unfolding networks (DUN) through a multi-scale and spatial-adaptive way. By integrating the flexible gradient descent and informative proximal mapping, we unfold the iterative PGD algorithm into a trainable DNN. Extensive experiments on various image restoration tasks demonstrate the superiority of our method in terms of state-of-the-art performance, interpretability, and generalizability. The source code is available at \href{https://github.com/MC-E/Deep-Generalized-Unfolding-Networks-for-Image-Restoration}{github.com/MC-E/DGUNet}.
\end{abstract}
\vspace{-10pt}

%%%%%%%%% BODY TEXT
\section{Introduction}
Image restoration (IR) aims to recover the high-quality image $\mathbf{x}$ from its degraded measurement $\mathbf{y}$. The degradation process is generally defined as:
\begin{equation}
\mathbf{y}=\mathbf{A}\mathbf{x}+\mathbf{n},
\label{deg}
\end{equation}
where $\mathbf{A}$ is the degradation matrix, and $\mathbf{n}$ represents the additive noise. It is typically an ill-posed problem. According to $\mathbf{A}$, IR can be categorized into many subtasks, \textit{e.g.}, image denoising, deblurring, deraining, compressive sensing. In the past few decades, IR has been extensively studied, leading to three main active research topics, \textit{i.e.}, model-based methods
% ~\cite{BM3D,nonlocalmean}
, deep learning methods
% ~\cite{dncnn,pnp2,ffdnet}
, and hybrid methods.
% ~\cite{dun2,dun3,pnp1,pnp2}.
% \vspace{-20pt}

\begin{figure}[t]
    \centering
    \includegraphics[width=0.95\linewidth]{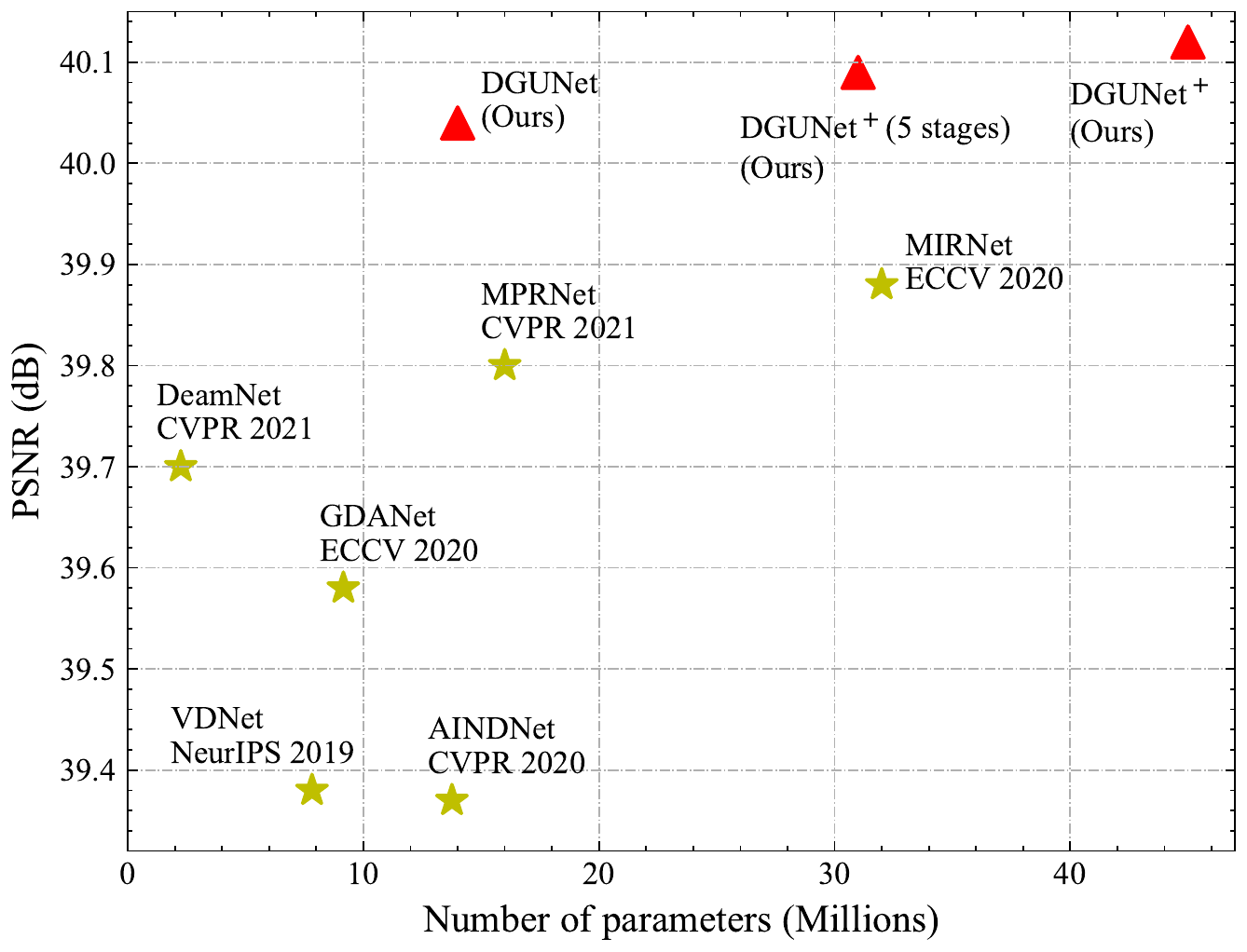}
    \caption{Real image denoising performance (y-axis) of our DGUNet and some recent denoisers (VDNet~\cite{vdnet}, GDANet~\cite{danet+}, AINDNet~\cite{aindnet}, MIRNet~\cite{mirnet}, DeamNet~\cite{deamnet}, MPRNet~\cite{MPRNet}) under different parameter capacities (x-axis) on DND~\cite{dnd} dataset.}
    \label{fig:complexity}
    \vspace{-10pt}
\end{figure}

% Regarding IR as a Bayesian problem, most model-based methods~\cite{classic_1,classic_2,classic_3,classic_4,BM3D,nonlocalmean} 
Model-based methods (\textit{e.g.}, \cite{classic_1,classic_2,classic_3,classic_4,BM3D,nonlocalmean}) usually formulate IR as a Bayesian problem, solving Eq.~\eqref{deg} under a unified MAP (maximizing a posterior) framework:
\begin{equation}
% \small
    \hat{\mathbf{x}}=\mathop{\mathtt{argmax}}\limits_{\mathbf{x}}{\log}P(\mathbf{x}|\mathbf{y})=\mathop{\mathtt{argmax}}\limits_{\mathbf{x}}{\log}P(\mathbf{y}|\mathbf{x})+{\log}P(\mathbf{x}),
    \label{map}
\end{equation}
where ${\log}P(\mathbf{y}|\mathbf{x})$ and ${\log}P(\mathbf{x})$ represent the data fidelity and regularization terms, respectively. The data fidelity term is usually defined as an $\ell_2$ norm, expressing Eq.~\eqref{map} as the following energy function:
\begin{equation}
    \hat{\mathbf{x}}=\mathop{\mathtt{argmin}}\limits_{\mathbf{x}}\frac{1}{2}||\mathbf{y}-\mathbf{A}\mathbf{x}||_2^2+\lambda J(\mathbf{x}),
    \label{optim}
    \vspace{-5pt}
\end{equation}
where $\lambda$ is a hyper-parameter to weight the regularization term $J(\mathbf{x})$. The data fidelity term guarantees the solution accords with degradation. The regulation term alleviates the ill-posed problem by enforcing desired property, which involves sophisticated priors, \textit{e.g.}, total variation~\cite{tv}, sparse representation~\cite{sp1,sp2,sp3}, low-rank~\cite{lowrank}, and self-similarity~\cite{nonlocalmean,BM3D}. However, the representation ability of handcrafted design is limited, leading to unstable results, and they are usually time-consuming in inference.

Recently, deep-learning IR \cite{arcnn,dncnn,ffdnet,pnp2} has achieved impressive success, as they can learn strong priors from large-scale datasets. Up to now, numerous function units have been proposed. \cite{memnet} proposed a memory strategy to broadcast useful information in different layers. \cite{cbdnet,vdnet,unprocessingnet,aindnet,MPRNet} utilized hourglass-shaped architectures to explore multi-scale features. 
Some non-local methods~\cite{nlrn,cola,dagl} were proposed to enlarge the receptive field. Although the promising performance and fast inference, the black box design makes it hard to analyze the role of different components, and performance gains are often attributed to stacking new modules at the price of increased model complexity.

To combine interpretability and adaptivity, some hybrid methods proposed integrating deep networks into classic optimization algorithms. For instance, deep plug-and-play (PNP) methods~\cite{pnp5,pnp1,pnp2,pnp3,pnp4} integrated pre-trained CNN denoiser, as the prior, into iterative optimization frameworks for different IR tasks. Unfortunately, they usually suffer from time-consuming inference. Recently, deep unfolding networks (DUN)\cite{dun2,dun3,pnpj,deamnet} proposed optimizing all parameters end-to-end, delivering better performance and faster inference. However, due to the interpretable design, most of them require known degradation processes to derive the solution. Nevertheless, the degradation processes in real-world applications are complicated and unassured with signal-dependent and spatially variant distribution. Thus, most DUN methods make handcrafted degradation assumptions~\cite{pnpj} or explicitly provide the network with degradation factors~\cite{dun2} to deal with pre-defined image degradation problems. In addition, since traditional model-based methods output an image in each iteration, the corresponding DUN has to adopt the input and output of each stage as an image. This inherent design inevitably results in feature-to-image information distortion. Such information loss in DUN has little notice in existing works.

To rectify the above issues and bridge the gap between model-based methods and deep learning methods, we propose a deep generalized unfolding network (DGUNet). On the one hand, our method has good interpretability as model-based methods by formulating the model design via a Proximal Gradient Descent (PGD) algorithm. On the other hand, similar to deep learning methods, our method is trained end-to-end with an unhindered feature pathway and can be easily applied to complex and real-world applications. To achieve this, we first integrate a gradient estimation strategy to the gradient descent step of the PGD algorithm to predict the gradient in degradation-unknown cases. We then design inter-stage information pathways to compensate for the intrinsic information loss in DUN. To summarize, this work has the following contributions:
\begin{itemize}
% \vspace{-10pt}
    \item The iterative optimization step of PGD algorithm is used to guide the mode design, leading to an end-to-end trainable and also interpretable model (DGUNet).
    \item Our DGUNet presents a general CNN-based implementation of DUN by combining a gradient estimation strategy into the PGD algorithm, enabling PGD to be easily applied to complex and real-world IR tasks.
    \item We design inter-stage information pathways in the DUN framework to broadcast multi-scale features in a spatial-adaptive normalization way, which rectifies the intrinsic information loss in most DUN methods.
    \item 
    % In addition to good interpretability, 
    Extensive experiments
    % on image deraining, deblurring, denoising and compressive sensing
    demonstrate that our method can solve general IR tasks with state-of-the-art performance (including \textbf{twelve} synthetic and real-world test sets) and attractive complexity (see Fig.~\ref{fig:complexity}).
\end{itemize}

\section{Related Works}
% Our proposed model is closely related to model-based algorithms, deep learning IR methods, and deep unfolding networks. Since in what follows, we give a brief review and focus on the specific methods most relevant to our own.

\subsection{Model-based Image Restoration Methods}
\label{opt_ista}
As mentioned previously, model-based methods~\cite{21-1,21-2,21-3,21-4} usually solve IR in a Bayesian perspective, which is formulated into a MAP optimization problem as Eq.~\eqref{optim}, containing a data fidelity term and a regularization term. HQS~\cite{hq}, ADMM~\cite{admm} and PGD~\cite{ista} are commonly used optimization algorithms. These methods usually decoupled the data fidelity term and regularization term of the objective function, resulting in an iterative scheme consisting of alternately solving a data subproblem and a prior subproblem. For instance, \cite{hqs_gmm} integrated Gaussian Mixture prior to HQS. In \cite{md1}, Heide \textit{et al.} used an alternative to ADMM and HQS to decouple the data term and prior term. \cite{md2} plugged class-specific Gaussian mixture denoiser into ADMM to solve image deblurring and compressive sensing.  

\begin{figure*}[t]
% \small
    \centering
    \includegraphics[width=1\linewidth]{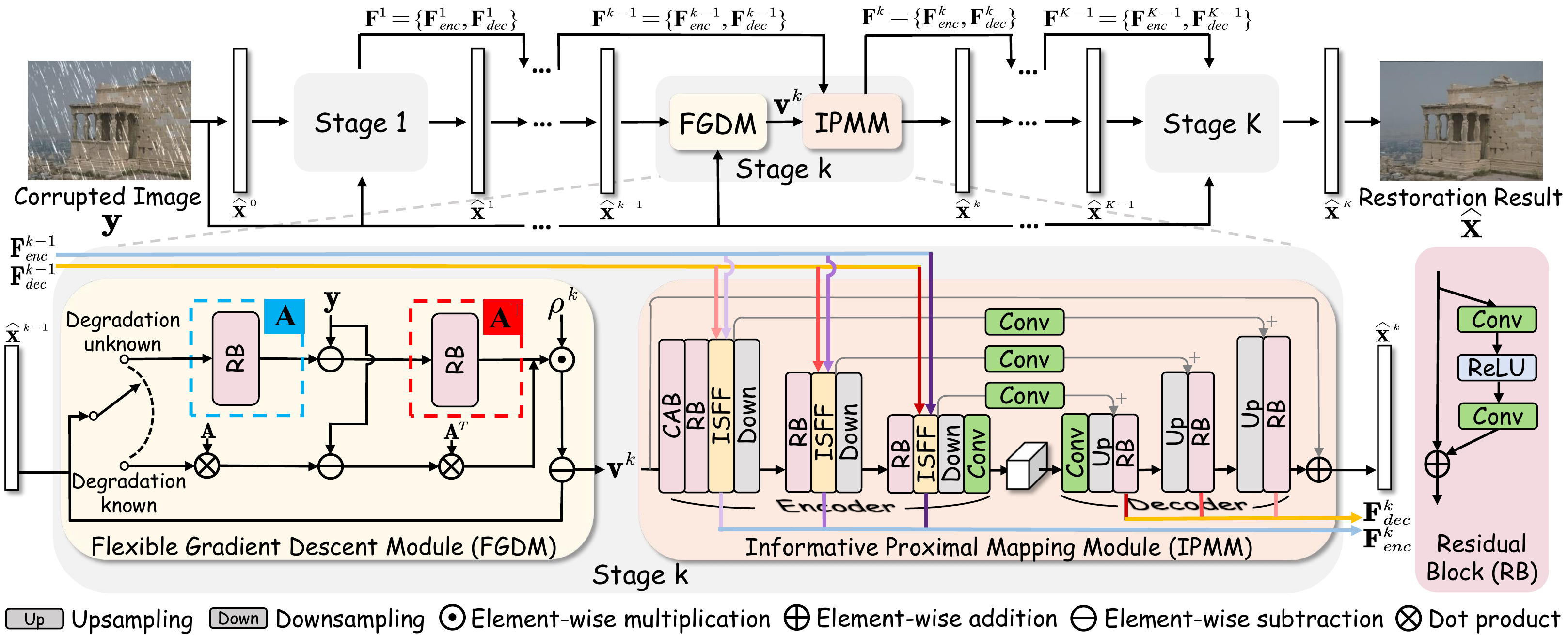}
    % \vspace{-5pt}
    \caption{Illustration of our proposed deep generalized unfolding network (DGUNet). We present the overall architecture in the first row, mainly composed of several stages. Each stage corresponds to an iteration in the PGD algorithm. The second row presents the detailed design of each stage, containing a flexible gradient descent module (FGDM) and an informative proximal mapping module (IPMM). 
    }
    \label{network}
    \vspace{-10pt}
\end{figure*}

% In this paper, we propose a more general CNN-based implementation of PGD for general image inverse problems.
% needed in second column of first page if using \IEEEpubid
%\IEEEpubidadjcol

\subsection{Deep Learning Image Restoration Methods} 
Motivated by the great success of deep neural networks (DNN), DNN-based methods have been widely used in low-level image processing tasks. \cite{arcnn,srcnn} are early attempts at applying convolutional neural networks (CNN) for IR. Subsequently, Zhang \textit{et al.} proposed DnCNN~\cite{dncnn}, significantly improving the restoration performance by residual learning. 
% Since DnCNN, several novel function units have been proposed. 
\cite{memnet} proposed dense-connected memory block to collect useful information from preceding layers. 
% To rectify the optimization problems of very deep networks for image restoration,
\cite{rcan,rnan} proposed stacking residual blocks (RB) without Batch Normalization~\cite{bn} to extend the network depth.
% and removing the normalization layer from the network. 
% \cite{nlrn,cola,dagl} explored the non-local self-similarity in IR.
% Realizing the importance of larger receptive field, dilated convolution~\cite{dilated,dilated2} and non-local networks~\cite{nlrn,n3net,cola,ipt} are often applied. 
In addition to estimating the clean image with a fixed scale, some hourglass-shaped networks~\cite{cbdnet,vdnet,unprocessingnet,aindnet,MPRNet} were proposed to explore multi-scale feature maps for IR.

\subsection{Deep Unfolding Networks} 
The main idea of deep unfolding networks (DUN) is that conventional iterative optimization algorithms can be implemented equivalently by a stack of recurrent DNN blocks. Such correspondence was originally applied in deep plug-and-play (PNP) methods~\cite{pnp5,pnp0,pnp1,pnp2,pnp3}, which utilize trained denoiser to implicitly express the regularization term $J(\mathbf{x})$ as a denoising problem. Inspired by PNP, DUN methods are trained in an end-to-end manner by jointly optimizing trainable denoisers in specific tasks. For instance, \cite{pnpj} jointly optimized an UNet as the proximal mapping in the ADMM~\cite{admm} algorithm. Nevertheless, its network structure is closely related to the handcrafted degradation assumptions to deal with pre-defined image degradation. \cite{dun2} used a ResUNet to replace the proximal mapping in the HQS~\cite{hq} algorithm. However, the degradation process is also manually designed, and its network requires scale factor, blur kernel, and noise level as additional inputs, causing the performance to depend largely on the accuracy of provided degradation factors. \cite{22-1,dun3,madun} solved compressive sensing by PGD algorithm~\cite{ista} with known degradation process. Moreover, most DUN methods are beset by information loss due to the feature-to-image transformation at the end of each stage. Though the skip connections in \cite{mogdun} benefit the information transfer, its implementation remains primitive, \textit{e.g.}, feature fusion is performed by concatenation on a single decoder layer of the proximal mapping module. 
% Thus, the generalization and restoration performance of DUN methods still have room for improvement.

\section{Methodology}
In this section, we first briefly review the traditional Proximal Gradient Descent (PGD) algorithm and then elaborate on our proposed DGUNet.

\subsection{Traditional Proximal Gradient Descent}
Technically, the PGD algorithm approximatively expresses Eq.~\eqref{optim} as an iterative convergence problem through the following iterative function:
\begin{equation}
\label{pgd}
    \textcolor{blue}{\hat{\mathbf{x}}^k = \mathop{\mathtt{argmin}}\limits_{{\mathbf{x}}}\frac{1}{2}||{\mathbf{x}}-}\textcolor{red}{(\hat{\mathbf{x}}^{k-1}-\rho \nabla g(\hat{\mathbf{x}}^{k-1}))}\textcolor{blue}{||_2^2+\lambda J({\mathbf{x}})},
\end{equation}
where $\hat{\mathbf{x}}^k$ refers to the output of the $k$-$th$ iteration, and $g(\cdot)$ represents the data fidelity term in Eq.~\eqref{optim}. $\nabla$ is the differential operator, weighted by the step size $\rho$. Mathematically, the \textcolor{red}{red} part of the above function is a gradient descent operation, and the \textcolor{blue}{blue} part can be solved by the proximal operator $\mathtt{prox}_{\lambda, J}$. Thus, it leads to two subproblems, \textit{i.e.}, gradient descent (Eq.~\eqref{ista1}) and proximal mapping (Eq.~\eqref{ista2}): 
\begin{subequations}
\begin{equation}
    \textcolor{red}{\mathbf{v}^k = \hat{\mathbf{x}}^{k-1}-\rho \mathbf{A}^\top(\mathbf{A}\hat{\mathbf{x}}^{k-1}-\mathbf{y})},
    \label{ista1}
\end{equation}
\begin{equation}
    \textcolor{blue}{\hat{\mathbf{x}}^k = \mathtt{prox}_{\lambda, J}(\mathbf{v}^k)}.
    \label{ista2}
\end{equation}
\end{subequations}
The PGD algorithm iteratively updates $\mathbf{v}^k$ and $\hat{\mathbf{x}}^k$ until convergence. 
% More details are provided in the supplementary material. 
ISTA~\cite{ista} is a typical PGD-based algorithm in which the regulation term is defined as an $\ell_1$ norm, \textit{i.e.}, $J(\mathbf{x})=||\mathbf{x}||_1$. Thus, the proximal mapping in ISTA is derived as a soft thresholding function: $\mathtt{prox}_{\lambda, J}(\mathbf{v}^k)=\mathtt{sign}(\mathbf{v}^k)\mathtt{max}(0,|\mathbf{v}^k|-\lambda)$. However, the handcrafted $\ell_1$ regulation has limited representation abilities, and its application is restricted to a few degradation-known tasks (\textit{e.g.}, compressive sensing). Focusing on improving the traditional PGD algorithm, in this paper, we unfold it by deep neural networks with robust and generalized design. 
% More details will be given in the following parts. 

\subsection{Proposed Deep Generalized Unfolding Network}
The whole network architecture of our proposed DGUNet is presented in Fig.~\ref{network}, which is an unfolding framework of the Proximal Gradient Descent (PGD) algorithm based on deep neural networks (DNN). Our DGUNet is composed of several repeated stages. Each stage contains a flexible gradient descent module (FGDM) and an informative proximal mapping module (IPMM), corresponding to the gradient descent (Eq.~\eqref{ista1}) and proximal mapping (Eq.~\eqref{ista2}) in an iteration step of the PGD algorithm, respectively. The number of stages is set as seven by default, and they share the same parameters except for the first and last stages. 
% Our proposed DGUNet comprises seven stages by default and shares all parameters among different stages except the first and last stages. 
To further improve the model performance, we also present a plus version, dubbed as DGUNet$^+$, in which all stages are parameter-independent.
\textbf{Flexible Gradient Descent Module.} As illustrated in Eq.~\eqref{ista1}, the gradient descent step is trivial when the degradation matrix $\mathbf{A}$ is known.
% \textit{e.g.}, $\mathbf{A}$ is the sampling matrix in compressive sensing. 
However, $\mathbf{A}$ is unknown in some degradation problems, making the gradient calculation (\textit{i.e.}, $\mathbf{A}^\top(\mathbf{A}\hat{\mathbf{x}}^{k-1}-\mathbf{y})$) intractable. 
% The handcrafted degradation assumption is not a reasonable choice in some real-world applications, whose noisy signals are signal-dependent and spatially variant. 
In this context, we propose a flexible gradient descent module (FGDM), shown in the second row of Fig.~\ref{network}. It has two model settings to deal with the degradation known and unknown cases reasonably. 

In the case of $\mathbf{A}$ is known, we directly use the accurate $\mathbf{A}$ to calculate the gradient. To improve the robustness, we set the step size $\rho$ as a trainable parameter in each stage, leading to the following gradient descent operation:
\begin{equation}
    \mathbf{v}^k = \hat{\mathbf{x}}^{k-1}-\rho^k \mathbf{A}^\top(\mathbf{A}\hat{\mathbf{x}}^{k-1}-\mathbf{y}).
    \label{ista1_new}
\end{equation} If $\mathbf{A}$ is unknown, instead of making task-specific assumptions for different degradation problems, we adopt a data-driven strategy to predict the gradient. Technically, we utilize two independent residual blocks, dubbed as $\mathcal{F}_{\mathbf{A}}^k$ and $\mathcal{F}_{\mathbf{A}^\top}^k$, to simulate $\mathbf{A}$ and its transpose $\mathbf{A}^\top$ in the $k$-$th$ stage. The gradient is calculated as $\mathcal{F}_{\mathbf{A}^\top}^k(\mathcal{F}_{\mathbf{A}}^k(\hat{\mathbf{x}}^{k-1})-\mathbf{y})$.
% To improve the robustness, we set the step size $\rho$ of gradient descent as a trainable parameter in each stage. 
Thus without loss of interpretability, the gradient descent in our proposed DGUNet can be defined as the following function in the degradation-unknown cases. 
\begin{equation}
\mathbf{v}^k = \hat{\mathbf{x}}^{k-1}-\rho^k \mathcal{F}_{\mathbf{A}^\top}^k(\mathcal{F}_{\mathbf{A}}^k(\hat{\mathbf{x}}^{k-1})-\mathbf{y}).
\label{ista1_new2}
\end{equation}
 
\begin{figure}[t]
% \small
    \centering
    \includegraphics[width=0.98\linewidth]{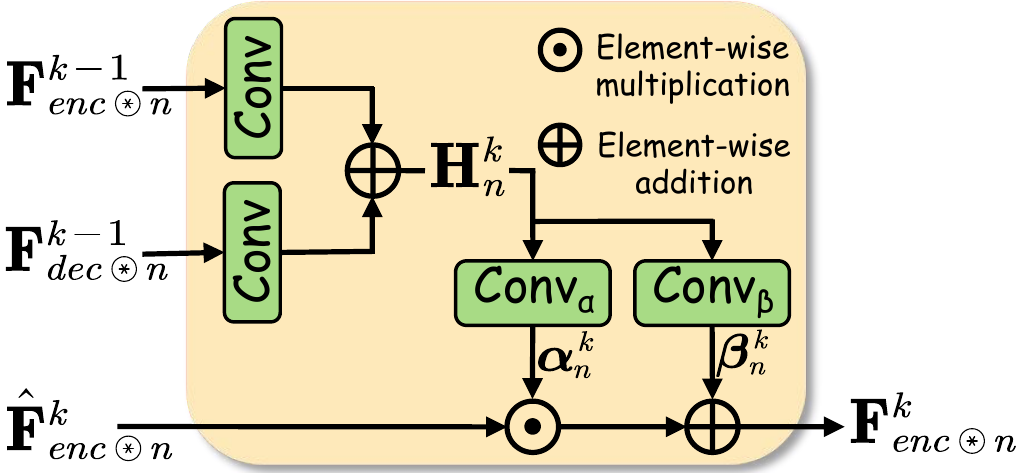}
    % \vspace{4pt}
    \caption{Illustration of our inter-stage feature fusion module (ISFF) at the $n$-$th$ scale in the $k$-$th$ stage. The encoder and decoder features from the previous stage are fused to the current stage in a spatial-adaptive normalization manner.}
    \vspace{-8pt}
    \label{feature_int}
\end{figure}

\begin{table*}[h]
\centering
\caption{
% Image deraining results on five datasets. 
Quantitative results (PSNR and SSIM) of image deraining. The best and second-best scores are \textbf{highlighted} and \underline{underlined}. 
% Our DGUNet and DGUNet$^+$ outperform the previous top-performing method MPRNet~\cite{MPRNet}, on average, by $0.28\ dB$ and $0.73\ dB$, respectively.
}
\vspace{-5pt}
\footnotesize
\begin{tabular}{l|cccccccccc|cc}
\shline
\multirow{2}{*}{\textbf{Method}}            & \multicolumn{2}{c}{Test100\cite{MPRNet}}     & \multicolumn{2}{c}{Rain100H\cite{yang2017deep}}    & \multicolumn{2}{c}{Rain100L\cite{yang2017deep}}    & \multicolumn{2}{c}{Test2800\cite{MPRNet}}    & \multicolumn{2}{c|}{Test1200\cite{MPRNet}}   & \multicolumn{2}{c}{Average}    \\
   & PSNR$\uparrow$           & SSIM$\uparrow$           & PSNR$\uparrow$           & SSIM$\uparrow$           & PSNR$\uparrow$           & SSIM$\uparrow$           & PSNR$\uparrow$           & SSIM$\uparrow$           & PSNR$\uparrow$           & SSIM$\uparrow$           & PSNR$\uparrow$           & SSIM$\uparrow$           \\ \shline
DerainNet~\cite{derainnet} & 22.77          & 0.810          & 14.92          & 0.592          & 27.03          & 0.884          & 24.31          & 0.861          & 23.38          & 0.835          & 22.48          & 0.796          \\
SEMI~\cite{semi}      & 22.35          & 0.788          & 16.56          & 0.486          & 25.03          & 0.842          & 24.43          & 0.782          & 26.05          & 0.822          & 22.88          & 0.744          \\
DIDMDN~\cite{didmdn}    & 22.56          & 0.818          & 17.35          & 0.524          & 25.23          & 0.741          & 28.13          & 0.867          & 29.65          & 0.901          & 24.58          & 0.770          \\
UMRL~\cite{umrl}      & 24.41          & 0.829          & 26.01          & 0.832          & 29.18          & 0.923          & 29.97          & 0.905          & 30.55          & 0.910          & 28.02          & 0.880          \\
RESCAN~\cite{rescan}    & 25.00          & 0.835          & 26.36          & 0.786          & 29.80          & 0.881          & 31.29          & 0.904          & 30.51          & 0.882          & 28.59          & 0.857          \\
PreNet~\cite{prenet}    & 24.81          & 0.851          & 26.77          & 0.858          & 32.44          & 0.950          & 31.75          & 0.916          & 31.36          & 0.911          & 29.42          & 0.897          \\
MSPFN~\cite{mspfn}     & 27.50          & 0.876          & 28.66          & 0.860          & 32.40          & 0.933          & 32.82          & 0.930          & 32.39          & 0.916          & 30.75          & 0.903          \\
MPRNet~\cite{MPRNet}    & 30.27    & 0.897    & 30.41    & 0.890    & 36.40    & 0.965    & 33.64    & 0.938    & 32.91    & \underline{0.916} & 32.73    & 0.921    \\
\hline
DGUNet (Ours)      & \underline{30.32} & \underline{0.899} & \underline{30.66} & \underline{0.891} & \underline{37.42} & \underline{0.969} & \underline{33.68} & \underline{0.938} & \textbf{33.23} & \textbf{0.920} & \underline{33.06} & \underline{0.923} \\
DGUNet$^{+}$ (Ours)      & \textbf{30.86} & \textbf{0.907} & \textbf{31.06} & \textbf{0.897} & \textbf{38.25} & \textbf{0.974} & \textbf{34.01} & \textbf{0.942} & \underline{33.08} & \underline{0.916}    & \textbf{33.46} & \textbf{0.927} \\ \shline
\end{tabular}
\label{tab:derainresult}
\vspace{-4pt}
\end{table*}

% \subsubsection{Informative Proximal Mapping Module}
% \label{pmm}
% Like most DUN methods~\cite{pnp1,pnp2,pnpj,dun2}, utilizing CNN-based denoiser to implicitly express the regulation term, we design an informative proximal mapping module (IPMM).
% It is well known that UNet is effective for image-to-image translation, and residual block (RB) is superior in increasing the model capacity. 
\textbf{Informative Proximal Mapping Module.} For the solution of Eq.~\eqref{ista2}, it is known that, from a Bayesian perspective, it actually corresponds to a denoising problem
% with the noise level $\sigma \triangleq\sqrt{1/ \lambda}$
~\cite{admm1,pnp5}.
% Regarding the proximal mapping as a denoising problem to implicitly express the regulation term, 
In this context, we design an informative proximal mapping module (IPMM), shown in the second row of Fig.~\ref{network}. Our IPMM is an hourglass-shaped architecture, consisting of an encoder and a decoder to utilize the multi-scale feature maps. Specifically, our IPMM begins with a channel attention block (CAB) to extract shallow features. We employ the residual block (RB) without Batch Normalization~\cite{bn} to extract features at three scales. Here, we utilize $\mathbf{F}_{enc}^{k} = \{\mathbf{F}_{enc\circledast n}^{k}\}_{n=1}^3$ and $\mathbf{F}_{dec}^{k}=\{\mathbf{F}_{dec\circledast n}^{k}\}_{n=1}^3$ to represent the encoder and decoder features extracted from the $k$-$th$ stage at the $n$-$th$ scale. 
% , and we follow the consensus~\cite{rnan,rcan} to remove the Batch Normalization~\cite{bn} from RB. 
% In each scale, there exists a skip connection pointing from encoder to decoder, and each connection contains a convolutional layer to refine the feature maps.
% The skip connection between encoder and decoder at each scale is equipped with a convolutional layer to refine the feature maps.
% At the end of each scale, we use bilinear sampling followed by a convolutional layer to perform downsampling and upsampling. 
In order to switch scales in our IPMM, we use $2\times 2$ max-pooling with stride 2 for downsampling, and we use bilinear upsampling followed by a convolutional layer for upsampling. Similar to many competitive denoisers, we add a global pathway from the input to the output, which encourages the network to bypass low-frequency information. At the end of IPMM, we utilize the supervised attention module (SAM) in \cite{MPRNet} to extract clean features and then inject them into the next stage through subspace projection~\cite{nbnet}.

Considering the intrinsic information loss in most DUN methods, we design inter-stage information pathways at each scale to broadcast useful information from encoder and decoder in different stages. For the illustration purpose, we use different colored lines to distinguish encoder and decoder information with different scales in Fig.~\ref{network}. To fuse the inter-stage information, we design an inter-stage feature fusion submodule (ISFF) at each scale in the encoder. Note that the inter-stage information can also be naturally propagated to the decoder due to the skip-connections between encoder and decoder. The detailed architecture of our ISFF is presented in Fig.~\ref{feature_int}, which is inspired by \cite{sft,sft2}. Concretely, at each scale, we transmit encoder and decoder features from the previous stage to the current stage.
% Considering the inter-stage information loss in most DUN methods, we design an inter-stage feature fusion submodule (ISFF) at each scale in the encoder to broadcast feature maps from the proximal mapping in different stages in a spatial-adaptive normalization manner. For illustration purpose, we utilize different colored lines to distinguish encoder and decoder inter-stage information with different scales in Fig.~\ref{network}. Note that due to the skip-connections between encoder and decoder, the inter-stage information can also be naturally propagated to the decoder. The detailed architecture of our ISFF is presented in Fig.~\ref{feature_int}, which is inspired by \cite{sft,sft2}. Concretely, we transmit the multi-scale features in the encoder and decoder from the previous stage to the current stage.
% The inter-stage features at each scale
They are first embedded by two independent $1\times 1$ convolutional layers and merged by the element-wise addition. In the $k$-$th$ stage, the fusion result at the $n$-$th$ scale is represented as $\mathbf{H}^{k}_{n}$. It is used to compute two affine parameters $\{ \bm{\alpha}^k_n,\bm{\beta}^k_n \}\in \mathbb{R}^{C\times H\times W}$ to transfer the intermediate output $\hat{\mathbf{F}}_{enc\circledast n}^{k}\in \mathbb{R}^{C\times H\times W}$ to an informative one $\mathbf{F}_{enc\circledast n}^{k}\in \mathbb{R}^{C\times H\times W}$, where $C$, $H$, and $W$ refer to the size of channel, height, and width, respectively. Mathematically, our proposed inter-stage feature fusion is defined as the following feature representation:
\begin{equation}
\begin{cases}
    &\hspace{-0.1cm} \mathbf{H}^{k-1}_n=\mathtt{Conv}(\mathbf{F}_{enc\circledast n}^{k-1})+\mathtt{Conv}(\mathbf{F}_{dec\circledast n}^{k-1})\\
    &\bm{\alpha}^k_n, \bm{\beta}^k_n=\mathtt{Conv}_{\alpha}(\mathbf{H}^{k-1}_n), \mathtt{Conv}_{\beta}(\mathbf{H}^{k-1}_n)\\
    &\mathbf{F}_{enc\circledast n}^{k} = \hat{\mathbf{F}}_{enc\circledast n}^{k}\bigodot \bm{\alpha}^k_n+\bm{\beta}^k_n.
    % &\hspace{-0.1cm} \mathbf{F}_{enc}^{k}=\mathbf{F}_{enc}^{k}\cdot Conv(\mathbf{H}^{k-1})+Conv(\mathbf{H}^{k-1})\\
    % &\hspace{0.7cm} =\mathbf{F}_{enc}^{k}\cdot \bm{\alpha}+\bm{\beta}
\end{cases}
\end{equation}

% Note that the feature fusion process 
The above feature fusion process is a standard spatial-adaptive normalization~\cite{sft}. Unlike conditional normalization methods \cite{bn,isn}, $\bm{\alpha}_n^k$ and $\bm{\beta}_n^k$ are not vectors but tensors with spatial dimensions. In this way, while the encoder and decoder obtain multi-scale features, feature maps at each scale can also have the refined memory of previous stages with the well-preserved spatial information, leading to an informative proximal mapping. For the illustration purpose, we utilize $\mathbf{F}^k$ to represent the set of multi-scale encoder and decoder features, \textit{i.e.}, $\mathbf{F}^k=\{\mathbf{F}_{enc}^{k},\ \mathbf{F}_{dec}^{k}\}$. Finally, our IPMM expresses Eq.~\eqref{ista2} as:
% \vspace{-4pt}
\begin{equation}
\hat{\mathbf{x}}^k, \mathbf{F}^{k} = \mathtt{prox}_{\bm{\theta}^k}(\mathbf{v}^k,\mathbf{F}^{k-1}),
\label{ista2_new}
\end{equation}
% \vspace{-4pt}
where $\bm{\theta}^k$ denotes the parameters of IPMM in the $k$-$the$ stage. In the light of the above discussion, we finally define the convergence process of our DGUNet in \textbf{Algorithm}~\ref{alg_1}.
\begin{algorithm}[tbh]
\caption{Proposed DGUNet}
\textbf{Initialization}:

    \hspace{0.01cm} (1) Initialize the iteration depth $k$=$0$ and ceiling~$K$;

    \hspace{0.01cm} (2) Initialize the input $\hat{\mathbf{x}}^{0} = \mathbf{y}$;
    
    \hspace{0.01cm} (3) Initialize the inter-stage feature $\mathbf{F}^{0} = None$;

\While{$k < K$}{

    \eIf{$\mathbf{A}$ is unknown}{
    Update $\mathbf{v}^{k+1}$ by Eq.~\eqref{ista1_new};
    %  $\mathbf{v}^{k+1} = \hat{\mathbf{x}}^{k}-\rho^k \mathcal{F}_{\mathbf{A}^\top}^{k+1}(\mathcal{F}_{\mathbf{A}}^{k+1}(\hat{\mathbf{x}}^{k})-\mathbf{y})$;
    }{
    Update $\mathbf{v}^{k+1}$ by Eq.~\eqref{ista1_new2};
    % \hspace{1.4cm} $\mathbf{v}^{k+1}$ is calculated by Eq.~\eqref{ista1};
    %  $\mathbf{v}^{k+1} = \hat{\mathbf{x}}^{k}-\rho^k\mathbf{A}^\top(\mathbf{A}\hat{\mathbf{x}}^{k}-\mathbf{y})$;
    }
    Update $\hat{\mathbf{x}}^{k+1}$ and $\mathbf{F}^{k+1}$ by Eq.~\eqref{ista2_new};
    % $\hat{\mathbf{x}}^{k+1},\ \mathbf{F}^{k+1} =$ $\mathtt{prox}_{\bm{\theta}^{k+1}}(\mathbf{v}^{k+1}$,\ $\mathbf{F}^{k})$;

    $k = k + 1$;}

% \hspace{0.2cm} \textbf{End while}

\textbf{Output:} $[\hat{\mathbf{x}}^1, \hat{\mathbf{x}}^2, \cdots ,\hat{\mathbf{x}}^K]$
\label{alg_1}
\end{algorithm}
\vspace{-5pt}

% \begin{algorithm}[tbh]
% \caption{Proposed DGUNet}
% $\bullet$ \textbf{Initialization}:

%     \hspace{0.5cm} (1) Initialize the iteration depth $k=0$;

%     \hspace{0.5cm} (2) Initialize the input as $\hat{\mathbf{x}}^{0} = \mathbf{y}$;
    
%     \hspace{0.5cm} (2) Initialize the inter-stage feature as $\mathbf{F}^{0} = None$;

% $\bullet$ \textbf{While} $k < K$ \textbf{do}:

%     \hspace{0.5cm} (1) if $\mathbf{A}$ is unknown:
    
%     \hspace{1.5cm}$\mathbf{v}^{k+1} = \hat{\mathbf{x}}^{k}-\rho^k \mathcal{F}_{\mathbf{A}^\top}^{k+1}(\mathcal{F}_{\mathbf{A}}^{k+1}(\hat{\mathbf{x}}^{k})-\mathbf{y})$;
    
%     \hspace{1cm} else:
    
%     % \hspace{1.4cm} $\mathbf{v}^{k+1}$ is calculated by Eq.~\eqref{ista1};
%     \hspace{1.4cm} $\mathbf{v}^{k+1} = \hat{\mathbf{x}}^{k}-\rho^k\mathbf{A}^\top(\mathbf{A}\hat{\mathbf{x}}^{k}-\mathbf{y})$;

%     \hspace{0.5cm} (2) $\hat{\mathbf{x}}^{k+1},\ \mathbf{F}^{k+1} =$ $\mathtt{prox}_{\bm{\theta}^{k+1}}(\mathbf{v}^{k+1}$,\ $\mathbf{F}^{k})$;

%     \hspace{0.5cm} (3) $k = k + 1$.

% \hspace{0.2cm} \textbf{End while}

% $\bullet$ \textbf{Output:} $[\hat{\mathbf{x}}^1, \hat{\mathbf{x}}^2, \cdots ,\hat{\mathbf{x}}^K]$
% \label{alg_1}
% \end{algorithm}

% \begin{algorithm}[tbh]
% \caption{Proposed DGUNet}
% \hspace*{0.02in} {\bf Input:} 
% input parameters A, B, C\\
% \hspace*{0.02in} {\bf Output:} 
% output result
% \begin{algorithmic}[1]
% \State some description 
% \For{condition} 
% 　　\State ...
% 　　\If{condition} 
% 　　　　\State ...
% 　　\Else
% 　　　　\State ...
% 　　\EndIf
% \EndFor
% \While{condition}
% 　　\State ...
% \EndWhile
% \State \Return result
% \end{algorithmic}
% \end{algorithm}

\subsection{Loss Function Design}
Without bells and whistles, we optimize our DGUNet and DGUNet$^+$ with the commonly used $\ell_2$ loss function, involving the output from all stages. Specifically, given the degraded measurement $\mathbf{y}$ and the ground-truth image $\mathbf{x}$, the goal of the training is defined as:
\begin{equation}
    \mathcal{L}(\mathbf{\Omega})=\sum_{k=1}^{K}\left\|\mathbf{x}-\hat{\mathbf{x}}^{k}\right\|^2_{2},
\end{equation}
where $K$ refers to the total number of stages, and $\hat{\mathbf{x}}^k$ represents the restoration result from the $k$-$th$ stage. $\mathbf{\Omega} = \{ \bm{\rho}^k,\mathcal{F}_{\mathbf{A}}^k(\cdot),\mathcal{F}_{\mathbf{A}^T}^k(\cdot),\bm{\theta}^k \}_{k=1}^K$ is the set of trainable parameters of our proposed DGUNet.

\begin{figure*}[h]
\centering
%-------------denoise result on Urban100 -------------
\begin{minipage}{0.27\linewidth}
\centering
\small
\includegraphics[width=1\columnwidth,height=3.2cm]{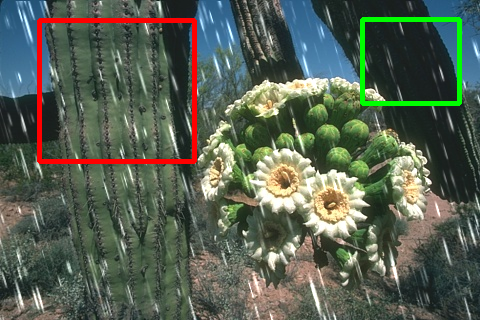}\\Rainy Image\\~\\
\end{minipage}
\begin{minipage}{0.135\linewidth}
\centering
\small
\includegraphics[width=1\columnwidth]{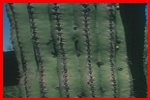}\\
\includegraphics[width=1\columnwidth]{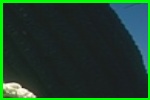}\\$PSNR$\\Ground Truth
\end{minipage}
\begin{minipage}{0.135\linewidth}
\centering
\small
\includegraphics[width=1\columnwidth]{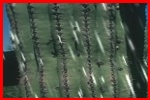}\\
\includegraphics[width=1\columnwidth]{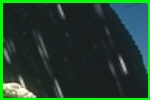}\\$22.88\ dB$\\Rainy
\end{minipage}
\begin{minipage}{0.135\linewidth}
\centering
\small
\includegraphics[width=1\columnwidth]{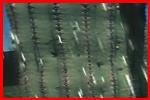}\\
\includegraphics[width=1\columnwidth]{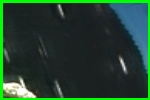}\\$23.41\ dB$\\DerainNet~\cite{derainnet}
\end{minipage}
\begin{minipage}{0.135\linewidth}
\centering
\small
\includegraphics[width=1\columnwidth]{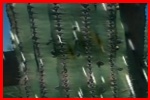}\\
\includegraphics[width=1\columnwidth]{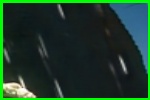}\\$23.86\ dB$\\SEMI~\cite{semi}
\end{minipage}
\begin{minipage}{0.135\linewidth}
\centering
\small
\includegraphics[width=1\columnwidth]{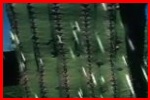}\\
\includegraphics[width=1\columnwidth]{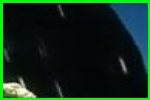}\\$23.17\ dB$\\DIDMDN~\cite{didmdn}
\end{minipage}

\begin{minipage}{0.135\linewidth}
\centering
\small
\includegraphics[width=1\columnwidth]{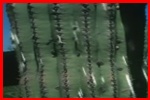}\\
\includegraphics[width=1\columnwidth]{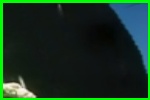}\\$26.77\  dB$\\UMRL~\cite{umrl}\vspace{-4pt}
\end{minipage}
\begin{minipage}{0.135\linewidth}
\centering
\small
\includegraphics[width=1\columnwidth]{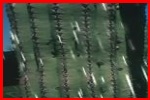}\\
\includegraphics[width=1\columnwidth]{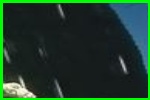}\\$25.50\ dB$\\RESCAN~\cite{rescan}\vspace{-4pt}
\end{minipage}
\begin{minipage}{0.135\linewidth}
\centering
\small
\includegraphics[width=1\columnwidth]{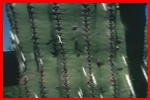}\\
\includegraphics[width=1\columnwidth]{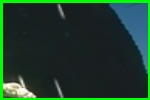}\\$26.07\ dB$\\PreNet~\cite{prenet}\vspace{-4pt}
\end{minipage}
\begin{minipage}{0.135\linewidth}
\centering
\small
\includegraphics[width=1\columnwidth]{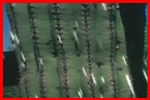}\\
\includegraphics[width=1\columnwidth]{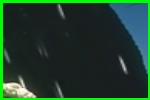}\\$25.13\ dB$\\MSPFN~\cite{mspfn}\vspace{-4pt}
\end{minipage}
\begin{minipage}{0.135\linewidth}
\centering
\small
\includegraphics[width=1\columnwidth]{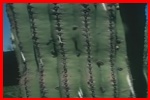}\\
\includegraphics[width=1\columnwidth]{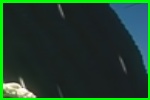}\\$30.54\ dB$\\MPRNet~\cite{MPRNet}\vspace{-4pt}
\end{minipage}
\begin{minipage}{0.135\linewidth}
\centering
\small
\includegraphics[width=1\columnwidth]{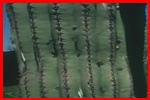}\\
\includegraphics[width=1\columnwidth]{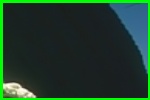}\\$32.25\ dB$\\DGUNet (ours)\vspace{-4pt}
\end{minipage}
\begin{minipage}{0.135\linewidth}
\centering
\small
\includegraphics[width=1\columnwidth]{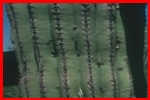}\\
\includegraphics[width=1\columnwidth]{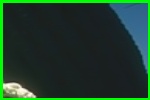}\\$32.58\ dB$\\DGUNet$^+$ (ours)\vspace{-4pt}
\end{minipage}
\caption{Visual comparison of image deraining. Our DGUNet and DGUNet$^+$ generate more natural details while removing raindrops.
% of image deraining application on widely used dataset Rain100L~\cite{yang2017deep}. 
% with DerainNet~\cite{derainnet}, SEMI~\cite{semi}, DIDMDN~\cite{didmdn}, UMRL~\cite{umrl}, RESCAN~\cite{rescan},  PreNet~\cite{prenet}, MSPFN~\cite{mspfn} and MPRNet~\cite{MPRNet}.
}
\label{fig:derainresult}
\vspace{-8pt}
\end{figure*}

\section{Experiments}
We apply our DGUNet to image deraining, deblurring, denoising, and compressive sensing. For each application, we train and evaluate our DGUNet with standard benchmarks and commonly used settings. The comparison is conducted with several recent methods. 
% We implement our DGUNet on MindSpore~\cite{mindspore} and other platforms. 
% Due to the page limitation, we present more results in supplementary materials.

\subsection{Training Details}
Our DGUNet is trained in an end-to-end manner. For image denoising, image deraining, and image deblurring tasks, we apply the same training strategy as MPRNet~\cite{MPRNet}. Specifically, we use Adam optimizer \cite{kingma2015adam}, with the initial learning rate being $2\times10^{-4}$. Considering the model depth, we utilize the warming up strategy~\cite{warm} to gradually improve the learning rate. The network is trained on $256\times256$ image patches, randomly cropped from training images. The batch size is set as $16$ for $4\times10^5$ iterations. In image compressive sensing, the network is trained on $32\times 32$ image patches, with the learning rate being $1\times10^{-4}$. The batch size is set as $128$ for $200$ epochs. 

The model training is performed on $2$ Nvidia Tesla V100 GPUs and can be completed within three days. For evaluation, we report standard metrics (PSNR and SSIM \cite{SSIM}). 

% \subsection{Performance Comparison}

\subsection{Image Deraining Results}
For image deraining, the training data is the same as MSPFN~\cite{mspfn} and MPRNet~\cite{MPRNet}. Specifically, we use 11,200 clean-rain image pairs in Rain14000~\cite{fu2017removing}, 1,800 image pairs in Rain1800~\cite{yang2017deep}, 700 image pairs in Rain800~\cite{zhang2019image} and 12 image pairs in Rain12~\cite{li2016rain} to train our model. For evaluation, five datasets, including Test2800~\cite{fu2017removing}, Test1200~\cite{didmdn}, Test100~\cite{zhang2019image}, Rain100H~\cite{yang2017deep} and Rain100L~\cite{yang2017deep} are utilized as the test sets. We compare our proposed DGUNet with eight competitive methods~\cite{derainnet,semi,didmdn,umrl,rescan,prenet,mspfn,MPRNet}. The quantitative comparison results are presented in Tab.~\ref{tab:derainresult}. One can see that our DGUNet and DGUNet$^+$ can outperform other methods on all test sets. Concretely, there are $1.85\ dB$, $0.65\ dB$, and $0.73\ dB$ gains compared with the recent MPRNet on Rain100L, Rain100H, and average of these five test sets, respectively. The visual comparison is presented in Fig.~\ref{fig:derainresult}, showing the better restoration result of our DGUNet and DGUNet$^+$. Especially compared with MSPFN and MPRNet, our method can remove inconspicuous rain lines better and produce more visually satisfying results with vivid details.

\begin{table}[t]
\centering
\caption{Image deblurring results on GoPro\cite{nah2017deep} and HIDE~\cite{shen2019human-aware}. The best and second-best scores are \textbf{highlighted} and \underline{underlined}. }
\footnotesize
\begin{tabular}{l|cc|cc}
% \hline
\shline
\multirow{2}{*}{Method}             & \multicolumn{2}{c|}{GoPro~\cite{nah2017deep}} & \multicolumn{2}{c}{HIDE~\cite{shen2019human-aware}} \\
       & PSNR$\uparrow$        & SSIM$\uparrow$        & PSNR$\uparrow$        & SSIM$\uparrow$       \\ \shline
Xu et al.~\cite{xu2013unnatural}    & 21.00       & 0.741       & -           & -          \\
Hyun et al.~\cite{hyun2013dynamic}  & 23.64       & 0.824       & -           & -          \\
Whyte et al.~\cite{whyte2012non-uniform} & 24.60       & 0.846       & -           & -          \\
Gong et al.~\cite{gong2017motion}  & 26.40       & 0.863       & -           & -          \\
DeblurGAN~\cite{deblurgan}    & 28.70       & 0.858       & 24.51       & 0.871      \\
Nah et al.~\cite{nah2017deep}  & 29.08       & 0.914       & 25.73       & 0.874      \\
Zhang et al.~\cite{zhang2018dynamic} & 29.19       & 0.931       & -           & -          \\
DeblurGAN-v2~\cite{deblurgan-v2} & 29.55       & 0.934       & 26.61       & 0.875      \\
SRN~\cite{srn}        & 30.26       & 0.934       & 28.36       & 0.915      \\
Shen et al.~\cite{shen2019human-aware} & -           & -           & 28.89       & 0.930      \\
Gao et al.~\cite{gao2019dynamic}  & 30.90       & 0.935       & 29.11       & 0.913      \\
DBGAN~\cite{DBGAN}       & 31.10       & 0.942       & 28.94       & 0.915      \\
MT-RNN~\cite{MTRNN}     & 31.15       & 0.945       & 29.15       & 0.918      \\
DMPHN~\cite{DMPH}      & 31.20       & 0.940       & 29.09       & 0.924      \\
Suin et al.~\cite{cvprdbl} & 31.85       & 0.948       & 29.98       & 0.930      \\
MPRNet~\cite{MPRNet}     & 32.66       & 0.959       & 30.96       &0.939      \\
\shline
DGUNet (Ours)         &      \underline{32.71}       &      \underline{0.960}       &      \underline{30.96}       &\underline{0.940}            \\
DGUNet$^+$ (Ours)        &      \textbf{33.17}       &      \textbf{0.963}       &      \textbf{31.40}       &        \textbf{0.944}    \\ 
\shline
\end{tabular}
\label{tab:deblurresult}
\vspace{-8pt}
\end{table}

\begin{figure*}[h]
\centering
\small 
\begin{minipage}[t]{0.118\linewidth}
\centering
\includegraphics[width=1\columnwidth,height=2cm]{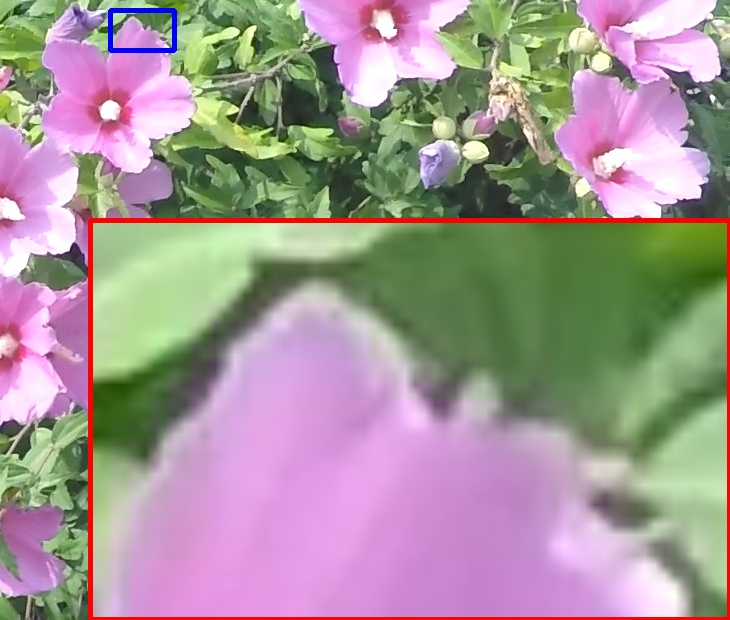}\\PSNR\vspace{1pt}
\includegraphics[width=1\columnwidth,height=2cm]{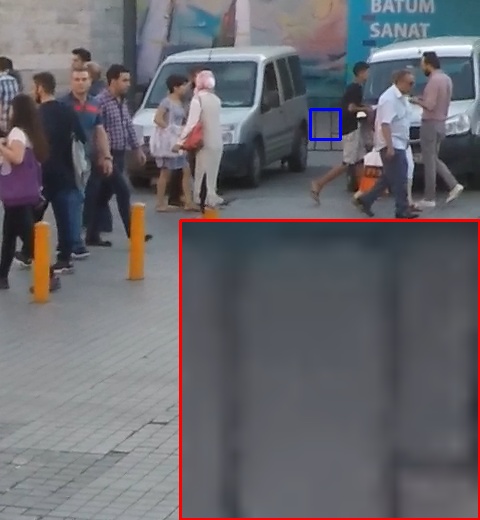}\\PSNR\\Ground Truth\vspace{-4pt}
\end{minipage}
\begin{minipage}[t]{0.118\linewidth}
\centering
\includegraphics[width=1\columnwidth,height=2cm]{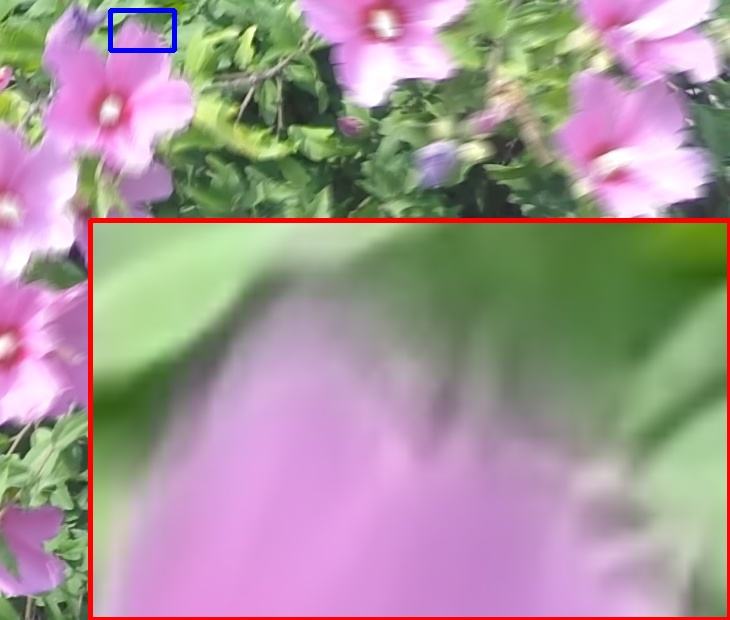}\\$26.51\ dB$\vspace{1pt}
\includegraphics[width=1\columnwidth,height=2cm]{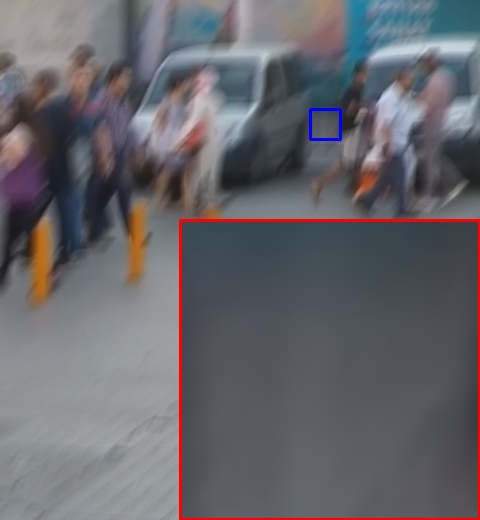}\\$25.43\ dB$\\Blurry Image\vspace{-4pt}
\end{minipage}
\begin{minipage}[t]{0.118\linewidth}
\centering
\includegraphics[width=1\columnwidth,height=2cm]{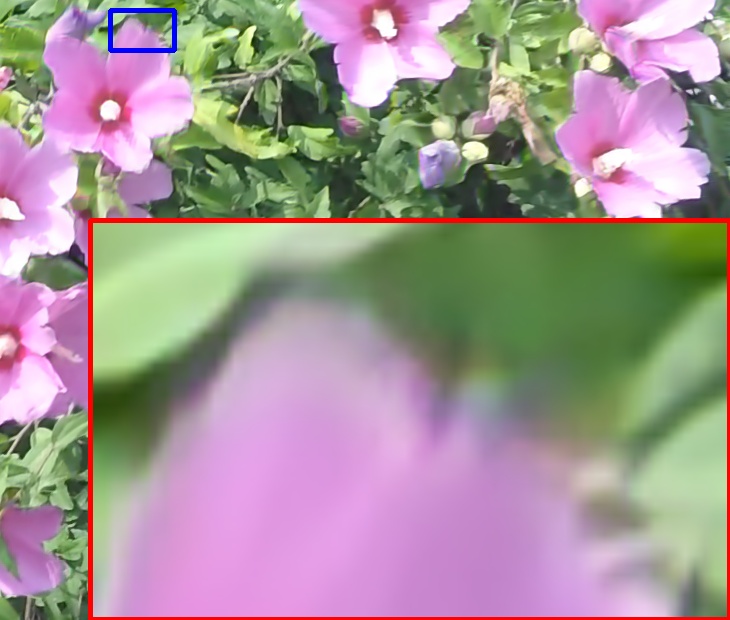}\\$29.46\ dB$\vspace{1pt}
\includegraphics[width=1\columnwidth,height=2cm]{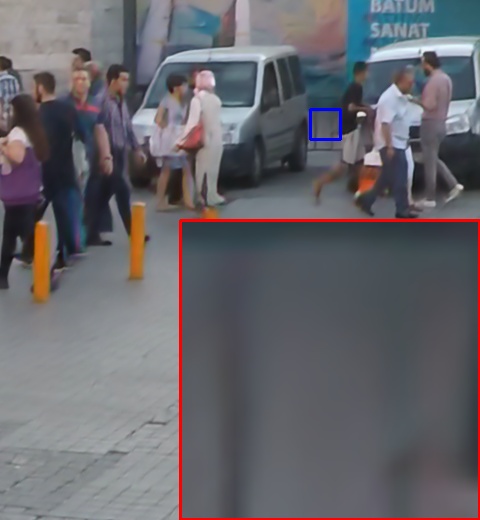}\\$32.87\ dB$\\Gao et al.~\cite{gao2019dynamic}\vspace{-4pt}
\end{minipage}
\begin{minipage}[t]{0.118\linewidth}
\centering
\includegraphics[width=1\columnwidth,height=2cm]{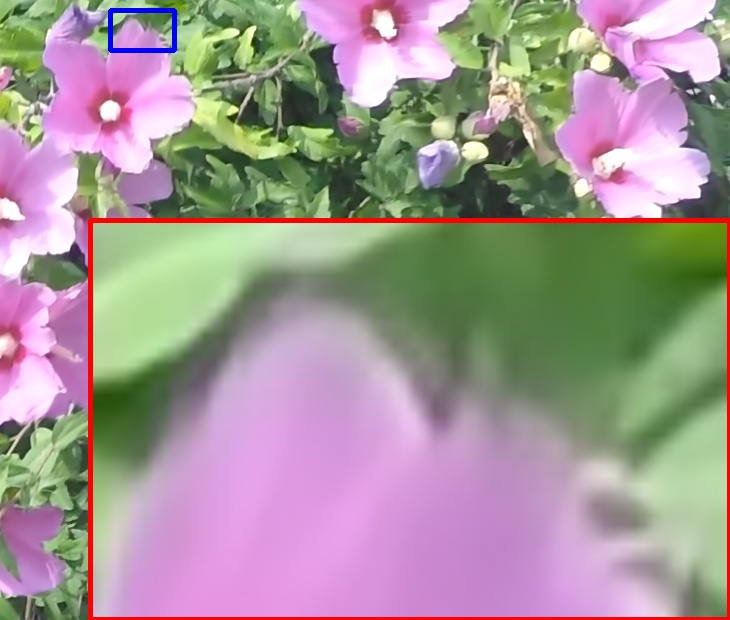}\\$29.77\ dB$\vspace{1pt}
\includegraphics[width=1\columnwidth,height=2cm]{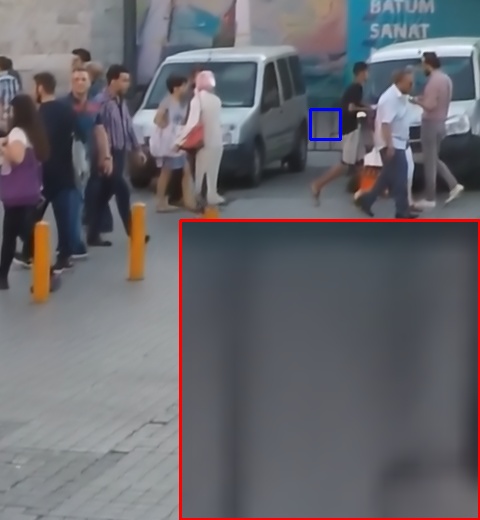}\\$33.48\ dB$\\DBGAN~\cite{DBGAN}\vspace{-4pt}
\end{minipage}
\begin{minipage}[t]{0.118\linewidth}
\centering
\includegraphics[width=1\columnwidth,height=2cm]{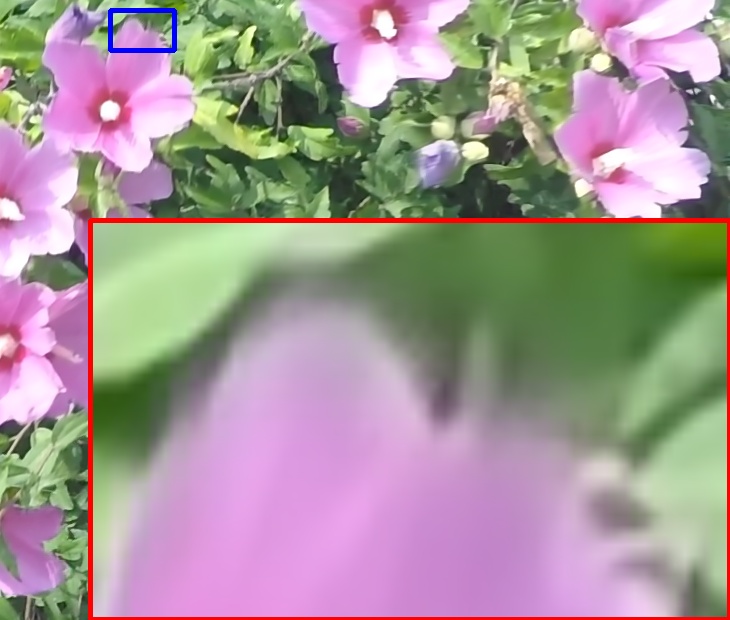}\\$29.71\ dB$\vspace{1pt}
\includegraphics[width=1\columnwidth,height=2cm]{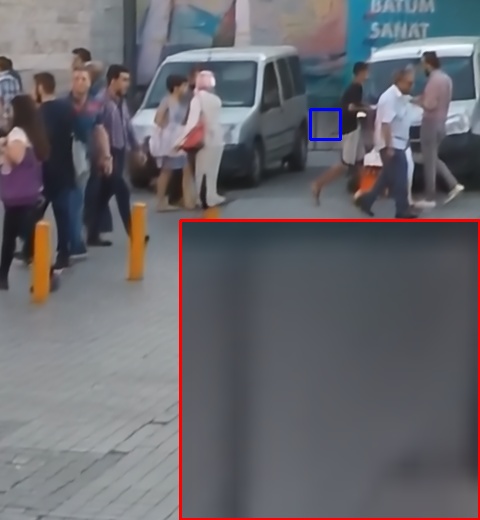}\\$32.88\ dB$\\MT-RNN~\cite{MTRNN}\vspace{-4pt}
\end{minipage}
\begin{minipage}[t]{0.118\linewidth}
\centering
\includegraphics[width=1\columnwidth,height=2cm]{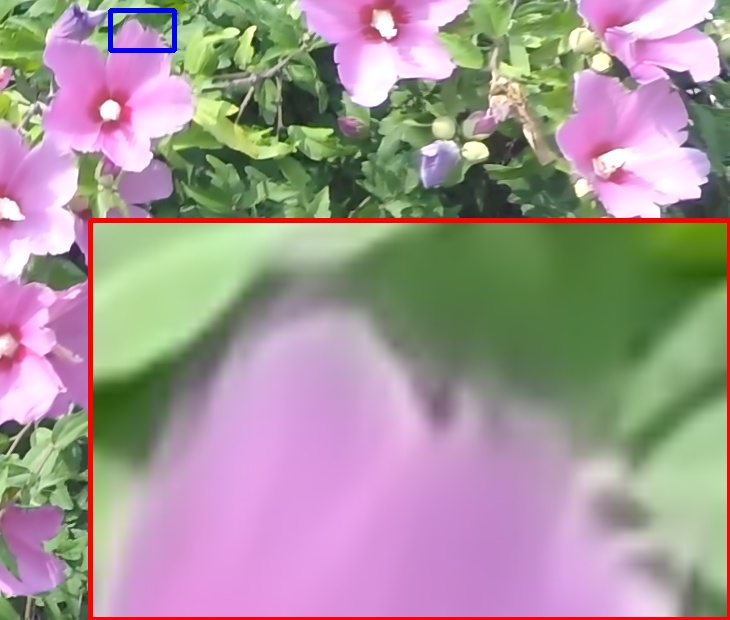}\\$30.58\ dB$\vspace{1pt}
\includegraphics[width=1\columnwidth,height=2cm]{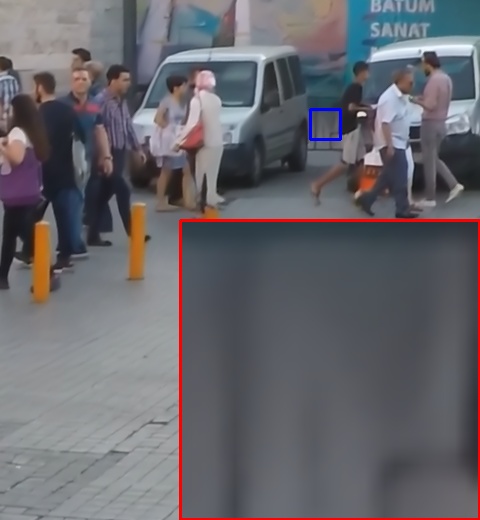}\\$35.13\ dB$\\MPRNet~\cite{MPRNet}\vspace{-4pt}
\end{minipage}
\begin{minipage}[t]{0.118\linewidth}
\centering
\includegraphics[width=1\columnwidth,height=2cm]{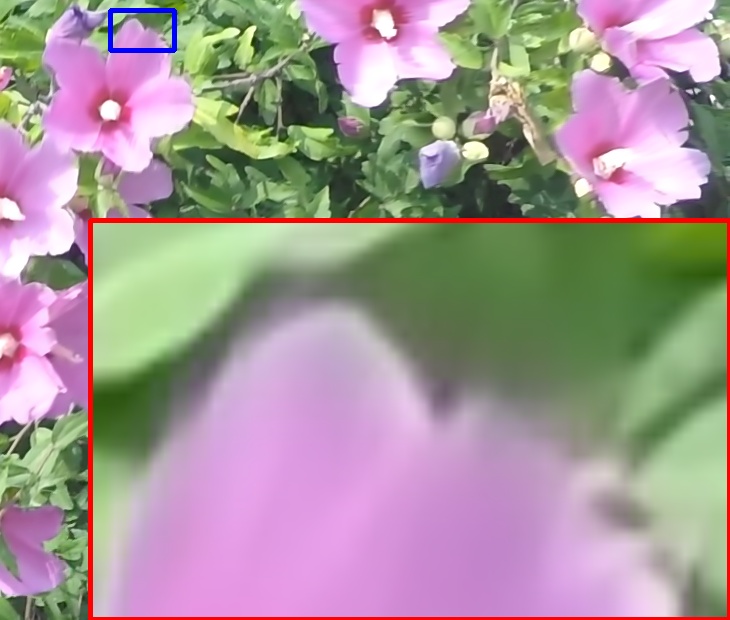}\\$30.67\ dB$\vspace{1pt}
\includegraphics[width=1\columnwidth,height=2cm]{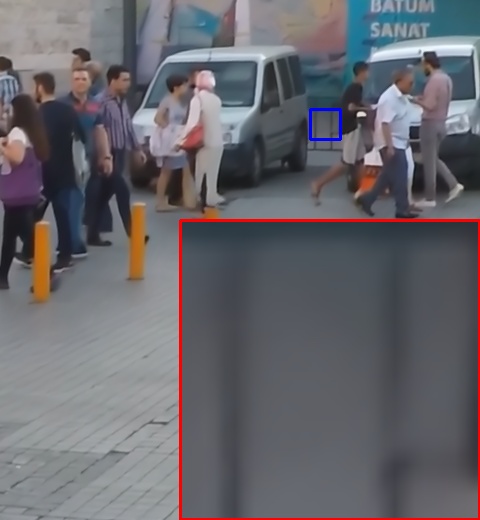}\\$35.37\ dB$\\DGUNet\vspace{-4pt}
\end{minipage}
\begin{minipage}[t]{0.118\linewidth}
\centering
\includegraphics[width=1\columnwidth,height=2cm]{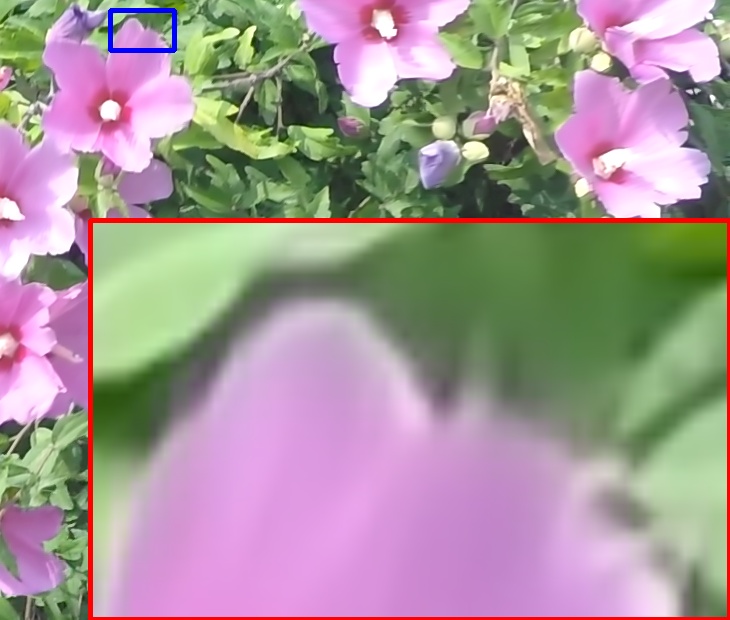}\\$30.89\ dB$\vspace{1pt}
\includegraphics[width=1\columnwidth,height=2cm]{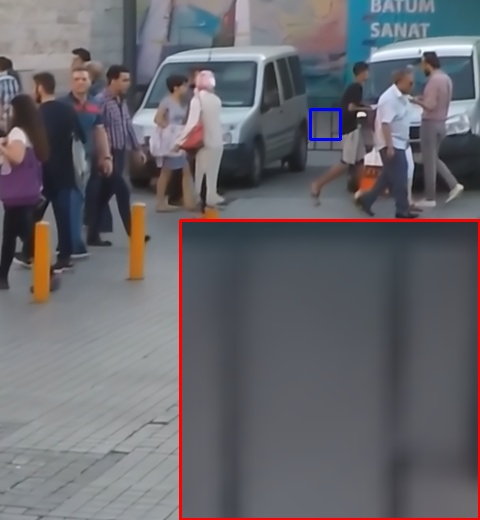}\\$35.97\ dB$\\DGUNet$^+$ \vspace{-4pt}
%\tiny{(Ours)}\vspace{-4pt}
\end{minipage}
\centering
\caption{Visual comparison of image deblurring. Our method produces sharper results that are visually closer to the ground truth.
% application on GoPro\cite{nah2017deep} valuation set. 
% with Gao et al.~\cite{gao2019dynamic}, DBGAN~\cite{DBGAN}, MT-RNN~\cite{MTRNN}, DMPHN~\cite{DMPH} and MPRNet~\cite{MPRNet}.
}
\label{fig:deblurresult} 
\end{figure*}

% ----------------------------------------------real noise--------------------
\begin{figure*}[h]
\centering
\small 
\begin{minipage}[t]{0.118\linewidth}
\centering
\includegraphics[width=1\columnwidth]{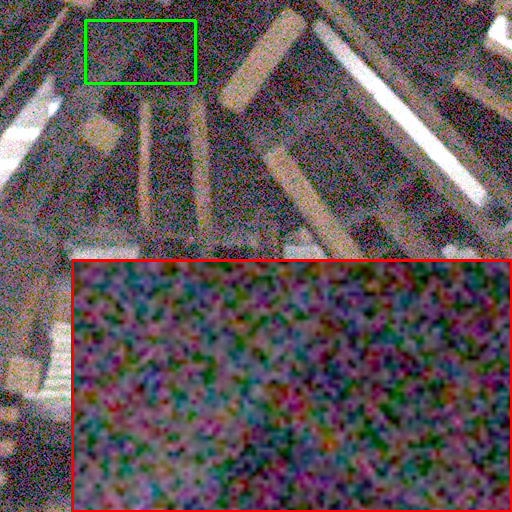}\\\vspace{1pt}
\includegraphics[width=1\columnwidth]{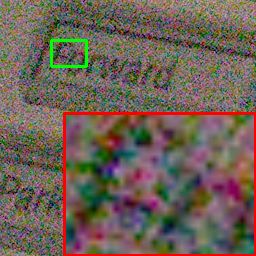}\\Noisy\vspace{-4pt}
\end{minipage}
\begin{minipage}[t]{0.118\linewidth}
\centering
\includegraphics[width=1\columnwidth]{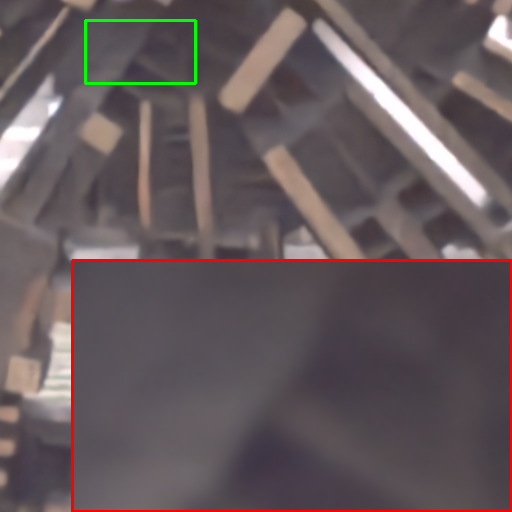}\\\vspace{1pt}
\includegraphics[width=1\columnwidth]{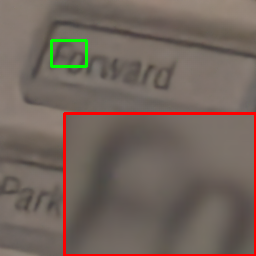}\\RIDNet \cite{ridnet}\vspace{-4pt}
\end{minipage}
\begin{minipage}[t]{0.118\linewidth}
\centering
\includegraphics[width=1\columnwidth]{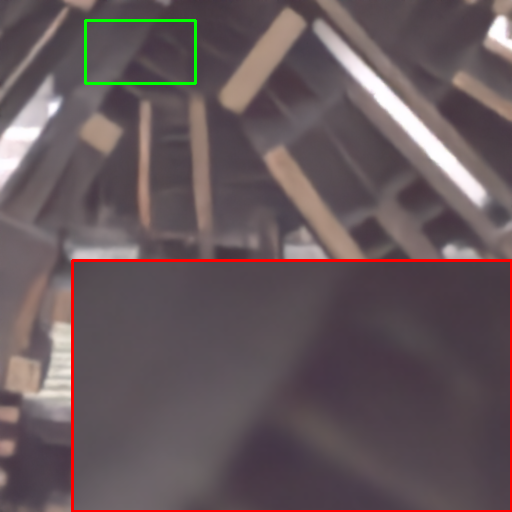}\\\vspace{1pt}
\includegraphics[width=1\columnwidth]{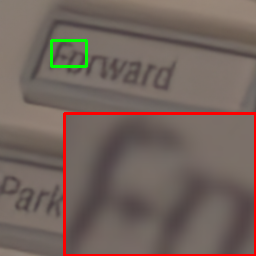}\\DANet \cite{danet+}\vspace{-4pt}
\end{minipage}
% \begin{minipage}[t]{0.118\linewidth}
% \centering
% \includegraphics[width=1\columnwidth]{images/real_noise/dnd/aindnet.png}\\\vspace{1pt}
% \includegraphics[width=1\columnwidth]{images/real_noise/sidd/aindnet.png}\\AINDNet \cite{aindnet}\vspace{-4pt}
% \end{minipage}
\begin{minipage}[t]{0.118\linewidth}
\centering
\includegraphics[width=1\columnwidth]{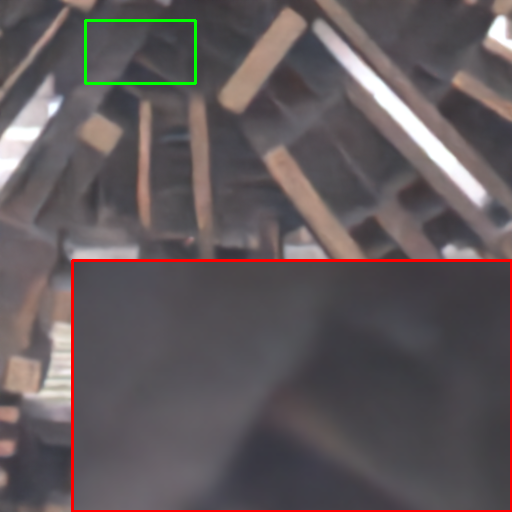}\\\vspace{1pt}
\includegraphics[width=1\columnwidth]{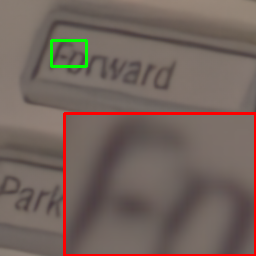}\\CycleISP \cite{cycleisp}\vspace{-4pt}
\end{minipage}
\begin{minipage}[t]{0.118\linewidth}
\centering
\includegraphics[width=1\columnwidth]{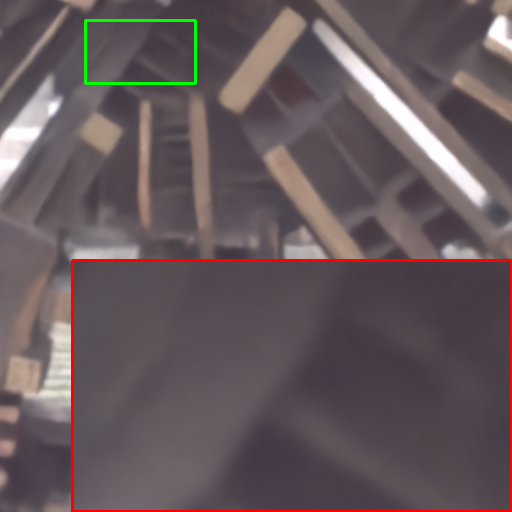}\\\vspace{1pt}
\includegraphics[width=1\columnwidth]{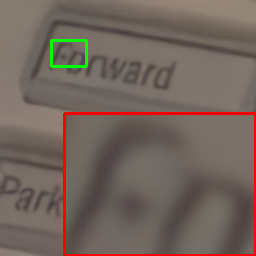}\\DeamNet \cite{deamnet}\vspace{-4pt}
\end{minipage}
\begin{minipage}[t]{0.118\linewidth}
\centering
\includegraphics[width=1\columnwidth]{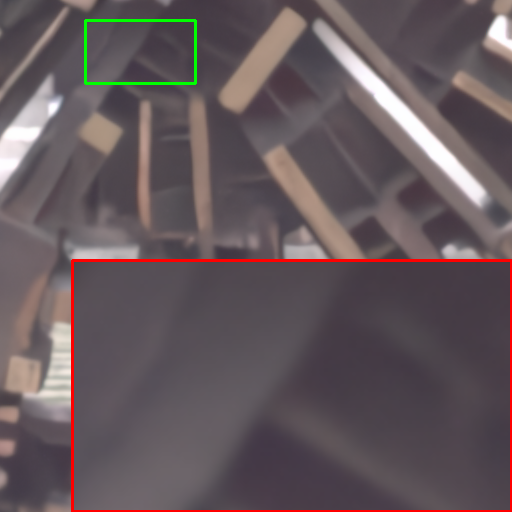}\\\vspace{1pt}
\includegraphics[width=1\columnwidth]{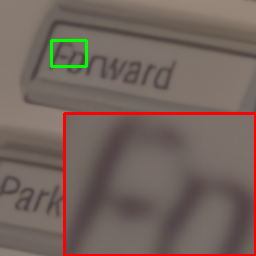}\\MPRNet \cite{MPRNet}\vspace{-4pt}
\end{minipage}
\begin{minipage}[t]{0.118\linewidth}
\centering
\includegraphics[width=1\columnwidth]{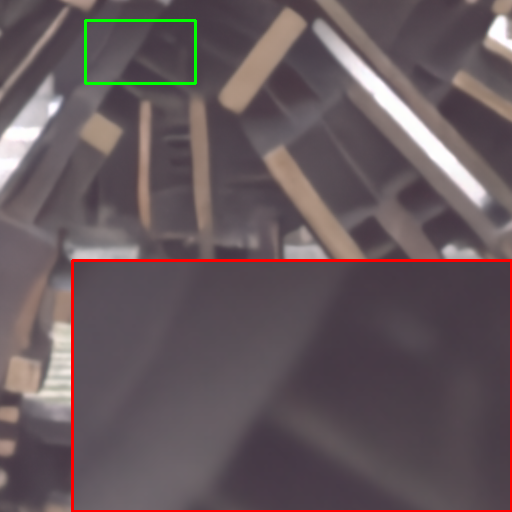}\\\vspace{1pt}
\includegraphics[width=1\columnwidth]{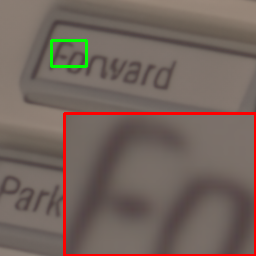}\\DGUNet \vspace{-4pt}
\end{minipage}
\begin{minipage}[t]{0.118\linewidth}
\centering
\includegraphics[width=1\columnwidth]{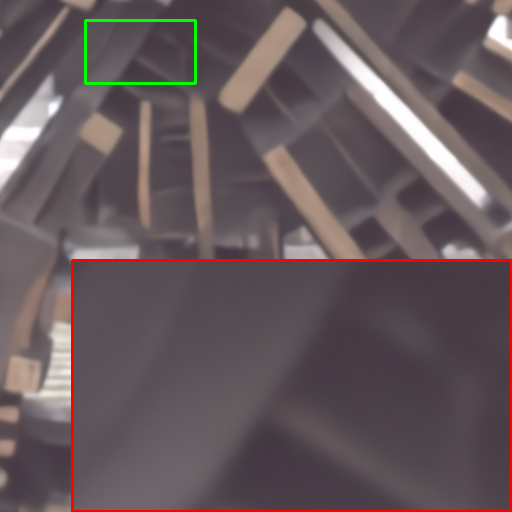}\\\vspace{1pt}
\includegraphics[width=1\columnwidth]{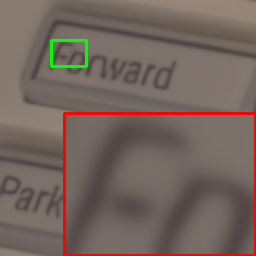}\\DGUNet$^+$ \vspace{-4pt}
% \tiny{(Ours)}\vspace{-4pt}
\end{minipage}
\centering
\caption{Visual comparison of real image denoising. Our method can remove unknown noise better while retaining satisfying details.
% These noisy images come from DND~\cite{dnd} dataset.
}
\label{fig:denoiseresult}
% \vspace{-5pt}
\end{figure*}

\subsection{Image Deblurring Results}
For image deblurring, similar to~\cite{rim2020real, DMPH, deblurgan, srn, MPRNet}, we train our model with $2,103$ image pairs from GoPro~\cite{nah2017deep} dataset and evaluate each method on the test sets from GoPro and HIDE~\cite{shen2019human-aware}, which contain $1,111$ and $2,025$ samples, respectively. Unlike pre-defined blur kernels, these two datasets are generated in real scenes, involving real-world degradation factors such as camera response function and human-aware motion blur. We compare our method with several competitive works~\cite{DBGAN,MTRNN,DMPH,cvprdbl} and the recent best algorithm MPRNet~\cite{MPRNet}. The quantitative evaluation is summarized in Tab.~\ref{tab:deblurresult}, presenting that our DGUNet and DGUNet$^+$ outperform other methods on these two test sets. Specifically, our DGUNet$^+$ outperforms MPRNet with $0.51\ dB$ and $0.44\ dB$ on GoPro and HIDE test sets. The visual comparison is shown in Fig.~\ref{fig:deblurresult}. Clearly, the results of our DGUNet and DGUNet$^+$ have higher visual quality, especially in recovering complex textures.
% , \textit{e.g.}, the petals and fences.

\begin{table}[t]
\centering
\caption{Image denoising results on SIDD~\cite{sidd} and DND~\cite{dnd}. The best and second-best scores are \textbf{highlighted} and \underline{underlined}. }
\footnotesize
\begin{tabular}{l|cc|cc}
\shline
\multirow{2}{*}{M\textbf{}ethod}             & \multicolumn{2}{c|}{SIDD~\cite{sidd}}                              & \multicolumn{2}{c}{DND~\cite{dnd}}                               \\
        & PSNR$\uparrow$                      & SSIM$\uparrow$                       & PSNR$\uparrow$                      & SSIM$\uparrow$                      \\ \shline
DnCNN~\cite{dncnn}        & 23.66                     & 0.583                      & 32.43                     & 0.790                     \\
MLP~\cite{mlp}          & 24.71                     & 0.641                      & 34.23                     & 0.833                     \\
BM3D~\cite{BM3D}         & 25.65                     & 0.685                      & 34.51                     & 0.851                     \\
CBDNet~\cite{cbdnet}       & 30.78                     & 0.801                      & 38.06                     & 0.942                     \\
RIDNet~\cite{ridnet}       & 38.71                     & 0.951                      & 39.26                     & 0.953                     \\
AINDNet~\cite{aindnet}      & 38.95                     & 0.952                      & 39.37                     & 0.951                     \\
VDN~\cite{vdnet}          & 39.28                     & 0.956                      & 39.38                     & 0.952                     \\
SADNet~\cite{sadnet}       & 39.46                     & 0.957                      & 39.59                     & 0.952                     \\
DANet+~\cite{danet+}       & 39.47                     & 0.957                      & 39.58                     & 0.955                     \\
MIRNet~\cite{mirnet}       & 39.72                    & 0.959                       & 39.88                     & 0.956                     \\
CycleISP~\cite{cycleisp}     & 39.52 & 0.957 & 39.56 & 0.956 \\
DeamNet~\cite{deamnet}  & 39.43 & 0.956 & 39.70 & 0.953\\
MPRNet~\cite{MPRNet}       & 39.71                     & 0.958                      & 39.80                     & 0.954                     \\ 
\shline
DGUNet (Ours) &       \underline{39.88}                    &        \underline{0.959}                    &  \underline{40.04}                         & \underline{0.956}                          \\
DGUNet$^+$ (Ours) &       \textbf{39.91}                    &       \textbf{0.960}                    & \textbf{40.12}                          &  \textbf{0.957}                         \\ \shline
\end{tabular}
\label{tab:denoiseresult}
\vspace{-8pt}
\end{table}

\begin{table*}[h]
\caption{Quantitative results of image compressive sensing. The best and second-best scores are \textbf{highlighted} and \underline{underlined}.}
\vspace{-4pt}
% \vspace{2pt}
\footnotesize
\centering
\begin{tabular}{c|c|c c c c c | c c}
\shline
Dataset & Ratio & ISTANet$^+$~\cite{dun3} & CSNet~\cite{csnet} & AdapRecon~\cite{adaptrecon} & OPINENet$^+$~\cite{opin} & AMPNet~\cite{ampnet} & DGUNet & DGUNet$^+$\\
\shline
\multirow{5}*{Set11} & $1\%$ & 17.42/0.4029 & 19.87/0.4977 & 19.63/0.4848 & 20.15/0.5340 & 20.04/0.5132& \underline{22.09}/\underline{0.6096} & \textbf{22.15}/\textbf{0.6114}\\
& $4\%$ & 21.32/0.6037 & 23.93/0.7338 & 23.87/0.7279 & 25.69/0.7920 & 24.64/0.7527& \textbf{26.84}/\textbf{0.8249}&\underline{26.83}/\underline{0.8230}\\
& $10\%$ & 26.64/0.8087 & 27.59/0.8575 & 27.39/0.8521 & 29.81/0.8884 & 28.84/0.8765& \textbf{31.07}/\textbf{0.9123}&\underline{30.93}/\underline{0.9088}\\
& $25\%$ & 32.59/0.9254 & 31.70/0.9274 & 31.75/0.9257 & 34.86/0.9509 & 34.42/0.9513 & \underline{36.11}/\underline{0.9611}&\textbf{36.18}/\textbf{0.9616}\\
& $50\%$ & 38.11/0.9707 & 37.19/0.9700 & 35.87/0.9625 & 40.17/0.9797 & 40.12/0.9818& \underline{41.22}/\underline{0.9836}&\textbf{41.24}/\textbf{0.9837}\\
\hline
\multirow{5}*{BSD68} & $1\%$ & 19.14/0.4158 & 21.91/0.4958 & 21.50/0.4825 & 22.11/0.5140 & 21.97/0.5086 & \underline{22.65}/\underline{0.5396} & \textbf{22.70}/\textbf{0.5406} \\
& $4\%$ & 22.17/0.5486 & 24.63/0.6564 & 24.30/0.6491 & 25.00/0.6825 & 25.40/0.6985& \textbf{25.55}/\textbf{0.7008}&\underline{25.45}/\underline{0.6987}\\
& $10\%$ & 25.32/0.7022 & 27.02/0.7864 & 26.72/0.7821 & 27.82/0.8045 & 27.41/0.8036 & \textbf{28.26}/\textbf{0.8193}&\underline{28.14}/\underline{0.8165}\\
& $25\%$ & 29.36/0.8525 & 30.22/0.8918 & 30.10/0.8901 & 31.51/0.9061 & 31.56/0.9121 & \underline{31.90}/\underline{0.9155} & \textbf{31.98}/\textbf{0.9158}\\
& $50\%$ & 34.04/0.9424 & 34.82/0.9590 & 33.60/0.9479 & 36.35/0.9660 & 36.64/0.9707& \underline{37.01}/\underline{0.9714}&\textbf{37.04}/\textbf{0.9718}\\
\shline
\end{tabular}
\label{tab:cs}
\end{table*}

\begin{figure*}[h]
\centering
\small 
\begin{minipage}[t]{0.118\linewidth}
\centering
\includegraphics[width=1\columnwidth]{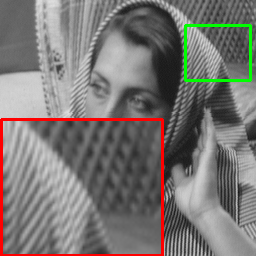}\\PSNR\\Ground Truth
\end{minipage}
\begin{minipage}[t]{0.118\linewidth}
\centering
\includegraphics[width=1\columnwidth]{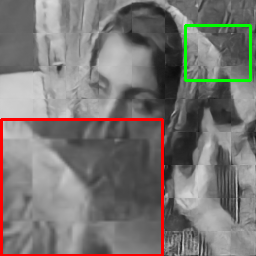}\\$23.51\ dB$\\ISTANet$^+$~\cite{dun3}
\end{minipage}
\begin{minipage}[t]{0.118\linewidth}
\centering
\includegraphics[width=1\columnwidth]{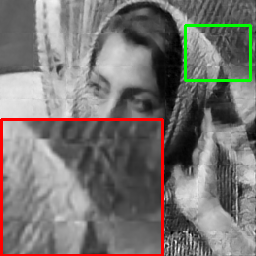}\\$23.06\ dB$\\AdapRecon~\cite{adaptrecon}
\end{minipage}
\begin{minipage}[t]{0.118\linewidth}
\centering
\includegraphics[width=1\columnwidth]{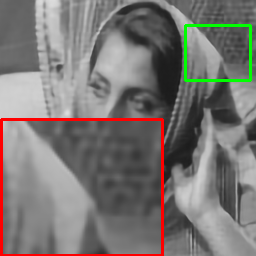}\\$24.41\ dB$\\CSNet~\cite{csnet}
\end{minipage}
\begin{minipage}[t]{0.118\linewidth}
\centering
\includegraphics[width=1\columnwidth]{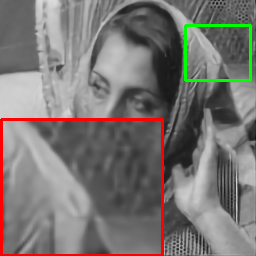}\\$24.73\ dB$\\OPINENet$^+$~\cite{opin}
\end{minipage}
\begin{minipage}[t]{0.118\linewidth}
\centering
\includegraphics[width=1\columnwidth]{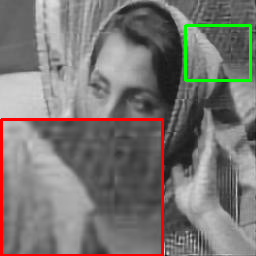}\\$24.56\ dB$\\AMPNet~\cite{ampnet}
\end{minipage}
\begin{minipage}[t]{0.118\linewidth}
\centering
\includegraphics[width=1\columnwidth]{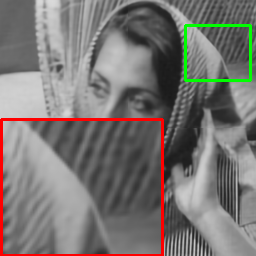}\\$26.13\ dB$\\DGUNet 
% (Ours)
\end{minipage}
\begin{minipage}[t]{0.118\linewidth}
\centering
\includegraphics[width=1\columnwidth]{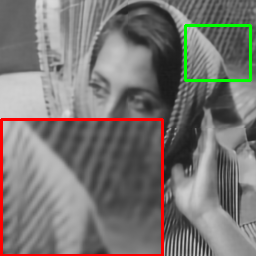}\\$26.41\ dB$\\DGUNet$^+$ 
% (Ours)
\end{minipage}
% \centering
\vspace{-6pt}
\caption{Visual comparison of compressive sensing with the CS ratio being $10\%$. Our method can produce results with higher quality.  
% application on Set11~\cite{nah2017deep} test set.
}
\label{fig:csresult} 
\vspace{-5pt}
\end{figure*}

% \begin{table*}[h]
%     \centering
%         \caption{Ablation study results (PSNR and SSIM) about some variants of our proposed method on Rain100H test set. 
%         }
%         \vspace{-4pt}
%         % \vspace{4pt}
% \small
%     \begin{tabular}{c|c c c c c c}
%     \shline
%         Mode & 7 Stages (DGUNet$^+$) & 5 Stages & 3 stages & w/o GDM & w/o ISFF & Weight sharing (DGUNet) \\
%     \hline
%          PSNR/SSIM & \textbf{31.13}/\textbf{0.900} & 30.43/0.889 & 29.67/0.878 & 30.51/0.887 & 30.09/0.882 & 30.68/0.895\\
%     \shline
%     \end{tabular}
%     \label{tab:abs}
% \end{table*}

% \begin{table}[h]
%     \centering
%         \caption{Ablation study of the number of stages and inter-stage feature fusion scales of our DGUNet on Rain100H test set. 
%         }
%         \vspace{-4pt}
%         % \vspace{4pt}
% \small
%     \begin{tabular}{c|c c c c| c c}
%     \shline
%     \multirow{2}*{Mode} &\multicolumn{4}{c|}{Number of stages} & \multicolumn{2}{c}{Number of scales} \\
%     \cline{2-7}
%          & 9 & 7 & 5 & 3 & 2 & 1\\
%     \hline
%          PSNR & \textbf{31.27} & 31.06 & 30.43 & 29.67 & 00.00 &00.00\\
%     \shline
%     \end{tabular}
%     \label{tab:abs1}
% \end{table}

\begin{table}[h]
    \centering
        \caption{Ablation study of the number of stages and number of feature fusion scales in our method on Rain100H test set. 
        }
        % \vspace{-6pt}
        % \vspace{4pt}
\small
    \begin{tabular}{c|c| c c c| c| c}
    \shline
    % \multirow{2}*{Mode} &\multicolumn{4}{c|}{Number of stages} & \multicolumn{2}{c}{Number of scales} \\
    % \cline{2-7}
    Stages & 9 &\multicolumn{3}{c|}{7} & 5 & 3\\
    \hline
    Scales & 3 & 3 & 2 & 1 & 3 &3\\
    \hline
         PSNR & \textbf{31.27} & 31.06  & 30.87 &30.68 & 30.43 & 29.67\\
    \shline
    \end{tabular}
    \label{tab:abs1}
    % \vspace{-4pt}
\end{table}

\begin{table}[h]
    \centering
        \caption{Ablation study of different components in our method.
        % on Rain100H test set. 
        }
        \vspace{-6pt}
        % \vspace{4pt}
\small
    \begin{tabular}{c|c c c c}
    \shline
        Mode & DGUNet$^+$ & w/o FGDM & w/o SP & w/o ISFF\\
    \hline
         PSNR & \textbf{31.06} & 30.51 & 30.92 & 30.09\\
    \shline
    \end{tabular}
    \label{tab:abs2}
    \vspace{-8pt}
\end{table}

\subsection{Image Denoising Results}
For this application, we train our DGUNet on the commonly used SIDD dataset~\cite{sidd}, which contains 320 degraded-clean image pairs corrupted by realistic noise with unknown distribution and noise levels. We evaluate each method on SIDD and DND~\cite{dnd} test sets. 
% $1,280$ validation patches from the SIDD and $1,000$ patches from the DND~\cite{dnd} validation sets. 
We compare our DGUNet with several recent methods~\cite{sadnet,danet+,cycleisp,deamnet,MPRNet} and report the evaluation results (PSNR and SSIM) in Tab.~\ref{tab:denoiseresult}. One can see that our method achieves the best performance on both SIDD and DND test sets. Specifically, our DGUNet$^+$ outperforms MPRNet with $0.32\ dB$ and $0.20\ dB$ on DND and SIDD test sets, respectively. The visual comparison is presented in Fig.~\ref{fig:denoiseresult}, including two samples from DND (the first row) and SIDD (the second row) test sets. Clearly, our method has good robustness to both high-intensity and low-intensity noise to recover the actual texture and structures, \textit{e.g.}, the pattern of wood and the edge of letters. 
% without compromising on the noise removal.

\subsection{Compressive Sensing Results}
For this application, we choose the widely used BSD400 dataset~\cite{BSD} as the training data and evaluate each method on Set11~\cite{Set11} and BSD68~\cite{BSD} test sets. Same as~\cite{adaptrecon,opin,ampnet}, for a given set of CS ratios $\{1\%, 4\%, 10\%, 25\%, 50\%\}$, we jointly optimize the sampling matrix with the whole network. Note that in the task of compressive sensing, the degradation matrix $\mathbf{A}$ is exactly known, \textit{i.e.}, the sampling matrix $\mathbf{\Phi}$. Thus, we directly use $\mathbf{\Phi}$ to calculate the gradient. The quantitative comparison is presented in Tab.~\ref{tab:cs}. One can see that our DGUNet and DGUNet$^+$ have obvious advantages over either classic methods~\cite{dun3,csnet} and recent top-performing methods~\cite{adaptrecon,opin,ampnet}, and the margin becomes more obvious at low CS ratios. For instance, there are $2\ dB$ gains compared with OPINENet$^+$~\cite{opin} on the Set11 test set, with the CS ratio being $1\%$. An interesting finding is that DGUNet performs better than DGUNet$^+$ in some cases. This is mainly because the training set is small and the large model can not be fully optimized. The visual comparison is presented in Fig.~\ref{fig:csresult}, showing that our method can recover more details and sharper edges than other methods.

\subsection{Ablation Study}
We present the ablation study in Tab.~\ref{tab:abs1} and Tab.~\ref{tab:abs2} to investigate the number of stages, number of feature fusion scales, and different components of our method. Experiments are conducted on image deraining and evaluated on Rain100H~\cite{yang2017deep}. In order to highlight the performance changes, all modifications are made on our DGUNet$^+$.

\textbf{Number of stages.} In this part, we explore the gains brought by the number of stages, including $9$, $7$, $5$, and $3$ stages. From Tab.~\ref{tab:abs1}, we can find that the performance increases with the number of stages, demonstrating the effectiveness of the iterative network design. By making a trade-off between performance and computational complexity, we employ seven stages in our DGUNet and DGUNet$^+$.

% \textbf{Number of inter-stage feature fusion scales.} 

\textbf{Inter-stage feature fusion.} As mentioned previously, to rectify the weakness of transforming a multi-channel feature map back to an image at the end of each stage, we introduce an inter-stage feature fusion module (ISFF).
% with multi-scales
% to compensate such information loss
% in a spatial-adaptive normalization manner. 
To demonstrate the effectiveness of ISFF, we remove it from our DGUNet$^+$, represented as ``w/o ISFF" in Tab.~\ref{tab:abs2}, and we study the performance gains of multi-scale feature fusion in Tab.~\ref{tab:abs1}. We can find that our ISFF has obvious gains, and the performance increases with the number of feature fusion scales. Additionally, we study the effectiveness of spatial-adaptive fusion by replacing it with the direct addition, represented as ``w/o SP" in Tab.~\ref{tab:abs2}. The performance degradation demonstrates the positive effect of this design.
% The experiment of multi-scales are presented in the Tab.~\ref{tab:abs1}, presenting that the performance increases with the number of scales.

\textbf{Gradient descent module.} Although our proposed DGUNet has good interpretability, the effect of such design also needs to be discussed carefully. In the experiment, we remove the Flexible Gradient Descent Module (FGDM) from our DGUNet$^+$, leading to an UNet-cascading structure. This variant is represented as ``w/o FGDM" in Tab.~\ref{tab:abs2}. Compared with DGUNet$^+$, there are $0.55\ dB$ degradations on the Rain100H. Thus, the result demonstrates that our interpretable design also has performance gains.

\section{Conclusion and Discussion}
In this paper, we propose a deep generalized unfolding network (DGUNet) for IR. We develop principles that aim to combine the merits of model-based methods and deep learning methods. To this end, we unfold the PGD optimization algorithm into a deep network and integrate a gradient estimation strategy into the gradient descent step, enabling it to be easily applied to complex and real-world applications. To compensate for the intrinsic information loss in DUN, we design inter-stage feature pathways that work with multiple scales and spatial-adaptive normalization. Extensive experiments on numerous IR tasks (including \textbf{twelve} synthetic and real-world test sets) demonstrate the superiority of our method in terms of state-of-the-art performance, interpretability, and generalizability. Our future work will support DGUNet on MindSpore~\cite{mindspore} platform.
% , which is a new deep learning computing framework.

% \textbf{Acknowledgments.} This work was sponsored by CAAI-Huawei Mindspore Open Fund~\cite{mindspore}.
% Extensive experiments on image denoising, deblurring, deraining, and compressive sensing demonstrate that our proposed model can achieve state-of-the-art performance on general IR tasks with good interpretability.

% Our DGUNet can be viewed as a general CNN-based implementation of the PGD algorithm by integrating a data-driven gradient estimator into the gradient descent step. In addition, we design an inter-stage feature fusion module that works with spatial-adaptive normalization to compensate for the intrinsic information loss in most DUN methods. Extensive experiments on image denoising, image deblurring, image deraining, and compressive sensing demonstrate that our proposed model can achieve state-of-the-art performance on general IR tasks with good interpretability. 
% and attractive complexity. 

% \textbf{Limitations.} 
% Though our method delivers outstanding performance on various IR tasks, the model parameters still have room for further compression. Our work focuses on producing higher signal fidelity, whereas its performance is unknown in scenes where perceptual quality is significantly important. These limitations warrant further research.

% \newpage
%%%%%%%%% REFERENCES
{\small
\bibliographystyle{ieee_fullname}
\bibliography{egbib}
}

\end{document}